\documentclass[10pt,twocolumn,twoside,journal]{IEEEtran}

\usepackage{amssymb}
\usepackage{booktabs}
\usepackage{cite}
\usepackage{hyperref}
\usepackage{xcolor}
\usepackage{stfloats}

\usepackage{bbding}
\usepackage{pifont}
\usepackage{textcomp}
\ifCLASSINFOpdf
\usepackage[pdftex]{graphicx}
\usepackage{caption}
\usepackage{subcaption}

\else
\fi

\usepackage{float}
\usepackage{multirow}
\usepackage{amsmath}
\usepackage{cleveref}
\newcommand{\etal}{\textit{et al}.}

\newcommand*{\zk}[1]{\textcolor{black}{#1}}

\begin{document}
	%
	\title{ Exploring Motion Ambiguity and Alignment for High-Quality Video Frame Interpolation }
	%
	%
	%
	\author{
		Kun Zhou\textsuperscript{1,2},
		\quad Wenbo Li\textsuperscript{3},
		\quad Xiaoguang Han\textsuperscript{1},~\IEEEmembership{Member,~IEEE},
		\quad Jiangbo Lu\textsuperscript{2},~\IEEEmembership{Senior~Member,~IEEE}
		
		\quad ${^1}$The Chinese University of Hong Kong~(Shenzhen) $^{2}$SmartMore Corporation \\ 
		$^{3}$The Chinese University of Hong Kong\\
	}

	{
	}
	%

	
	
	\maketitle
		\begin{figure*}[t]
		\centering
		\includegraphics[width=18cm]{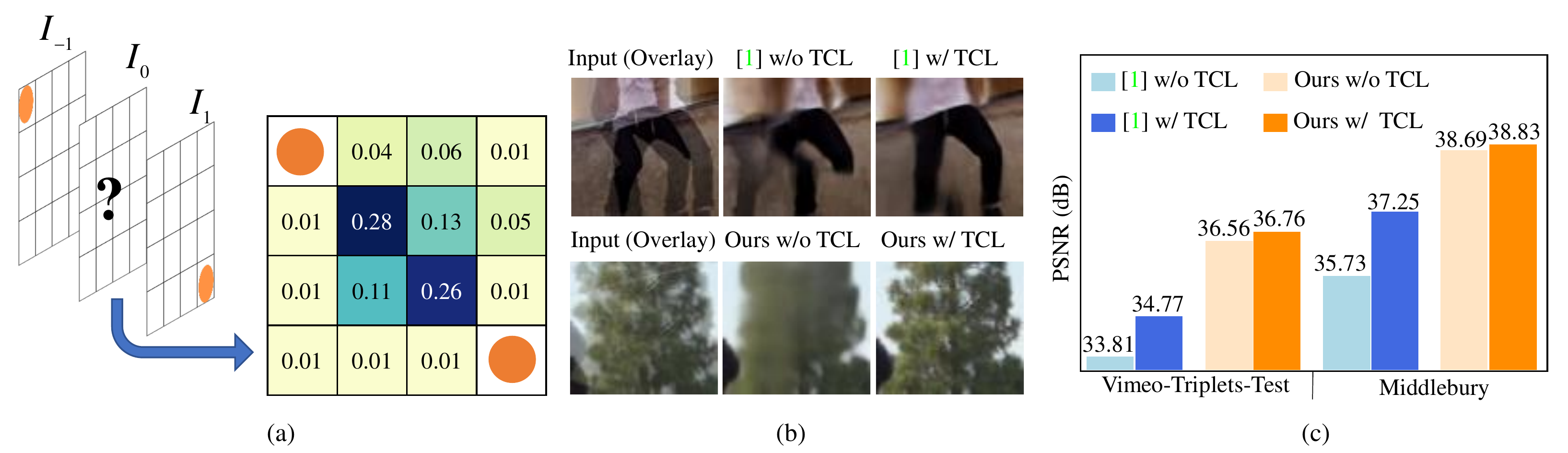} 
		
		\caption{Analysis of motion ambiguity in VFI. (a) User study of querying the location of a ball in the intermediate frame $I_0$ with the observed two input frames $\{I_{-1},I_1\}$. The results are visualized in a probability distribution map. (b) Visual comparison between SepConv~\cite{niklaus2017videosep} and our method with/without the proposed texture consistency loss (TCL). (c) Quantitative evaluation of the two methods with/without TCL loss on on Vimeo-Triplets~\cite{xue2019video} and Middlebury~\cite{baker2011database} benchmarks.
		} 
	\vspace{-0.15in}
		\label{fig:teasing}
	\end{figure*} %
	
%
	
	\begin{abstract}
		\label{sec:abstract}
For video frame interpolation~(VFI), existing deep-learning-based approaches strongly rely on the ground-truth (GT) intermediate frames, which sometimes ignore the non-unique nature of motion judging from the given adjacent frames. As a result, these methods tend to produce averaged solutions that are not clear enough. To alleviate this issue, we propose to relax the requirement of reconstructing an intermediate frame as close to the GT as possible. Towards this end, we develop a texture consistency loss (TCL) upon the assumption that the interpolated content should maintain similar structures with their counterparts in the given frames. Predictions satisfying this constraint are encouraged, though they may differ from the pre-defined GT. Without the bells and whistles, our plug-and-play TCL is capable of improving the performance of existing VFI frameworks. On the other hand, previous methods usually adopt the cost volume or correlation map to achieve more accurate image/feature warping. However, the $O(N^2)$~(\zk{$N$ refers to the pixel count}) computational complexity makes it infeasible for high-resolution cases. In this work, we design a simple, efficient ($O(N)$) yet powerful cross-scale pyramid alignment~(CSPA) module, where multi-scale information is highly exploited. Extensive experiments justify the efficiency and effectiveness of the proposed strategy.

Compared to state-of-the-art VFI algorithms, our method boosts the PSNR performance by 0.66dB on the Vimeo-Triplets dataset and 1.31dB on the Vimeo90K-7f dataset. In addition, our method is easily extended to the video frame extrapolation task. Surprisingly, our extrapolation model has achieved a 0.91dB PSNR gain over FLAVR under the same experimental setting, while being 2$\times$ times smaller in terms of the model size. At last, we show that our high-quality interpolated frames are also beneficial to the development of the video super-resolution task.

	\end{abstract}


	%
	\IEEEpeerreviewmaketitle
	

	\section{ Introduction}
	\label{sec:intro}

	\IEEEPARstart{V}{ideo} frame interpolation~(VFI) plays a critical role in computer vision with numerous applications, such as video editing and novel view synthesis. Unlike other vision tasks that heavily rely on human annotations, VFI benefits from the abundant off-the-shelf videos to generate high-quality training data. The recent years have witnessed the rapid development of VFI empowered by the success of deep neural networks. The popular approaches can be roughly divided into two categories: 1) optical-flow-based methods~\cite{huang2020rife,niklaus2020softmax,liu2017video,jiang2018super,NEURIPS2019_d045c59a,niklaus2018context,zhang2020flexible,liu2020enhanced,siyao2021deep,zhao2021ea,kubas2021fastrife,zhai2005low,jayashankar2019lap,xue2021bwin,chi2020all,xing2021flow} and  2) kernel-regression-based algorithms~\cite{niklaus2017videosep,cheng2020video,chen2021pdwn,lee2020adacof,bao2019memc,niklaus2017video,shi2021video}.
	
	The optical-flow-based methods typically warp the images/features based on a linear or quadratic motion model and then complete the interpolation by fusing the warped results. Nevertheless, it is not flexible enough to model the real-world motion under the linear or quadratic assumption, especially for cases with long-range correspondence or complex motion. Besides, occlusion reasoning is a challenging problem for pixel-wise optical flow estimation. Without the prerequisites above, the kernel-based methods handle the reasoning and aggregation in an implicit way, which adaptively aggregate neighboring pixels from the images/features to generate the target pixel. However, this line stands the chance of failing to tackle the high-resolution frame interpolation or large motion due to the limited receptive field. Thereafter, deformable convolutional networks, a variant of kernel-based methods, are adopted to aggregate the long-term correspondence~\cite{cheng2021multiple,chen2021pdwn,lee2020adacof}, achieving better performance. Despite many attempts, some challenging issues remain unresolved.

	First, the deep-learning-based VFI works focus on learning the predefined ground truth (GT) and ignore the inherent motion diversity across a sequence of frames. As illustrated in Fig.~\ref{fig:teasing}~(a), given the positions of a ball in frames ${I}_{-1}$ and ${I}_{1}$, we conduct a user study of choosing its most possible position in the intermediate frame ${I}_{0}$. The obtained probability distribution map clearly clarifies the phenomenon of motion ambiguity in VFI. Without considering this point, existing methods that adopt the pixel-wise L1 or L2 supervision possibly generate blurry results, as shown in Fig.~\ref{fig:teasing}~(b). To resolve this problem, we propose a novel texture consistency loss (TCL) that relaxes the rigid supervision of GT while ensuring texture consistency across adjacent frames. Specifically, for an estimated patch, apart from the predefined GT, we look for another texture-matched patch from the input frames as a pseudo label to jointly optimize the network. In this case, predictions satisfying the texture consistency are also encouraged. From the visualization comparison of SepConv~\cite{niklaus2017videosep} and our model with/without TCL~{\footnote{The four models are trained on Vimeo-Triplets~\cite{xue2019video} dataset.}} in Fig.~\ref{fig:teasing}~(b), we observe that the proposed TCL leads to clearer results. Besides, as shown in Fig.~\ref{fig:teasing}~(c), it is seen that our TCL  brings about considerable PSNR improvement on Vimeo-Triplets~\cite{xue2019video} and Middlebury~\cite{baker2011database} benchmarks for both two methods. More visual examples are available in our appendix~\ref{sec:AppenA}. 

	Second, the cross-scale aggregation during alignment is not fully exploited in VFI. For example, PDWN~\cite{chen2021pdwn} conducts an image-level warping using the gradually refined offsets. However, the single-level alignment may not take full advantage of the cross-scale information,  which has been proven useful in many low-level tasks~\cite{zhang2014cross,li2020mucan,li2021best}. In this work, we propose a novel cross-scale pyramid alignment~(CSPA) module, which performs bidirectional temporal alignment from low-resolution stages to higher ones. In each step, the previously aligned low-scale features are regarded as a guidance for the current-level warping. To aggregate the multi-scale information, we design an efficient fusion strategy rather than building the time-consuming cost volume or correlation map. Extensive quantitative and qualitative experiments verify the effectiveness and efficiency of the proposed method. 
	
	In a nutshell, our contributions are summarized as follows:
	\begin{itemize}
		
		
		%
		
	\item	\textit{\bf Texture consistency loss}: Inspired by the motion ambiguity in VFI, we design a novel texture consistency loss to allow the diversity of interpolated content, producing clearer results.
	
	\item \textit{\bf Cross-scale pyramid alignment}: The proposed alignment strategy utilizes the multi-scale information to conduct a more accurate and robust motion compensation while requiring few computational resources.
	\item \textit{\bf State-of-the-art performance}: The extensive experiments including single-frame and multi-frame interpolation have demonstrated the superior performance of the proposed algorithm. 
	
	\item 	\textit{\bf Extension to other tasks:} Based on the same architecture, our model is easily tailored to the video frame extrapolation task. Moreover, we take advantage of our well-trained model to generate high-quality intermediate frames, which can be naturally utilized in the video super-resolution task. We show that the synthesized images further bring the existing video super-resolution methods to new heights.
	\end{itemize}
	\begin{figure*}[t]
		\centering
		\includegraphics[width=18cm]{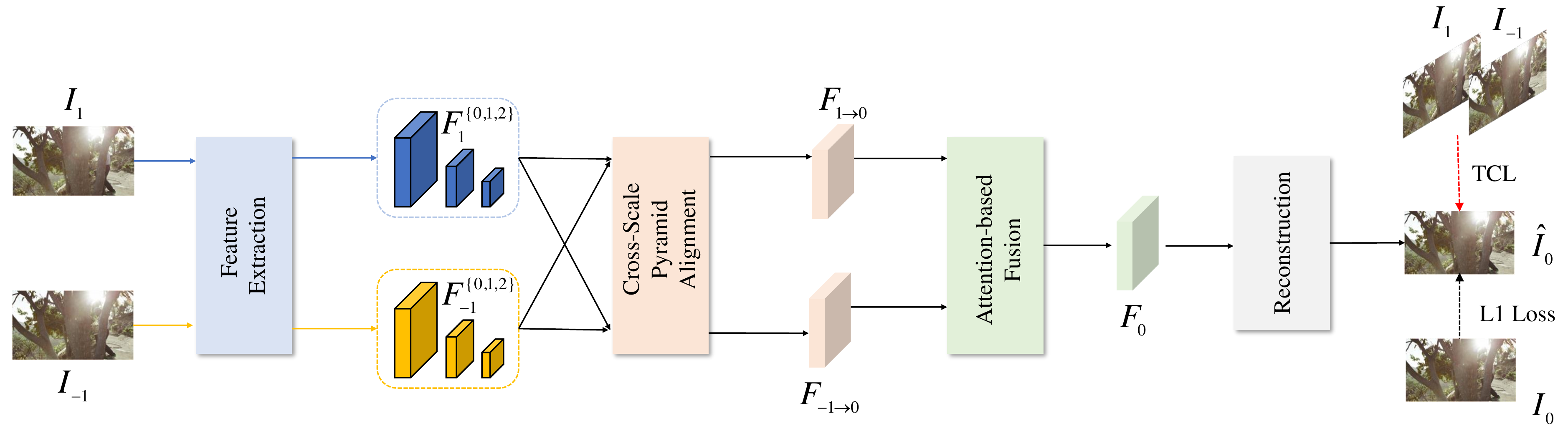} 
		
		\caption{Overview of the proposed VFI architecture. There are four components including a feature extraction module, a cross-scale pyramid alignment module, an attention-based fusion module, and a reconstruction module. In addition to the L1 loss for supervision, we propose a texture consistency loss (TCL) to encourage the diversity of objects' motion.
		} 
		\label{fig:framework}
	\end{figure*} %
	\section{Related Work}
	\label{sec:related}
	\subsection{Optical-Flow-Based Methods}
A large group of methods utilize optical flow to build pixel-wise correspondences, thereafter, they warp the given neighboring frames to the target frame. For example, TOFlow~\cite{xue2019video} designs a task-oriented optical flow module and achieves favorable results compared with approaches using off-the-shelf optical flow. Jiang \etal~\cite{jiang2018super} propose an end-to-end convolutional neural network to interpret the bidirectional motion based on the optical flow meanwhile reasoning the occlusions from a visibility map. With the help of the linear combination of bidirectional warped features, they can interpolate frames at an arbitrary time. In~\cite{niklaus2018context}, Niklaus and Liu present a context-aware frame interpolation approach by introducing additional warped deep features to provide rich contextual information. Huang \etal~\cite{huang2020rife} devise a lightweight sub-module named IFNet to predict the optical flow and train it in a supervised way. Choi \etal~\cite{choi2021high} propose a tridirectional motion estimation method to obtain more accurate optical flow fields. To resolve the conflict of mapping multiple pixels to the same target location in the forward mapping, Niklaus \etal~\cite{niklaus2020softmax} develop a differential softmax splatting method achieving a new state of the art. However, these optical flow-based methods generally have a poor performance when facing some challenging cases, such as large occlusion and complex motion.
	
	\subsection{Kernel-Regression-Based Methods}
	In addition to optical-flow-based methods above, learning adaptive gathering kernels~\cite{niklaus2017video,peleg2019net,cheng2021multiple,chen2021pdwn,lee2020adacof,niklaus2017videosep,NEURIPS2020_eaae339c} has also received intensive attention. Niklaus \etal~\cite{niklaus2017video} regard the video frame interpolation as a local convolution over the input frames. A U-Net is designed to regress a pair of kernels that are applied on the input frames to handle the alignments and occlusions simultaneously. To reduce the model parameters while maintaining a comparable receptive field, Niklaus \etal~\cite{niklaus2017videosep} propose another separable convolution network by combining two 1D kernels into a 2D adaptive kernel. However, both of these methods obtain limited performance when dealing with large displacements due to the restricted kernel size. To enlarge the receptive field, Peleg \etal~\cite{peleg2019net} develop a multi-scale feature extraction module to capture long-distance correspondences. Recently, deformable convolution networks~(DCN)~\cite{cheng2021multiple,chen2021pdwn,zhou2021revisiting,lee2020adacof} have shown a great success in the field of video frame interpolation. In PDWN~\cite{chen2021pdwn}, the authors design a pyramid deformable network to warp the contents of input frames to the target frame. While these kernel-based VFI approaches show high flexibility and good performance, they neglect the essential cross-scale information from input frames. In this work, we devise a cross-scale pyramid alignment to fuse multiple features in different resolutions, achieving better performance.

	\subsection{Beyond Linear Motion Model}
	
	Another line of studies focuses on investigating the physical motion from more input frames. For instance,  Yin \etal~\cite{NEURIPS2019_d045c59a} present a quadratic video interpolation method that takes the acceleration information into consideration, performing a better approximation of the complex motion in the real world. Later on, Zhang \etal~\cite{NEURIPS2020_9a118833} develop a well-generalized model to analyze the complex motion patterns, which further boosts the interpolation quality. Though more input frames can be used to better understand motion properties, this kind of methods need to carefully handle the abundant clues for reconstruction.
	
	\subsection{Census Transform}
	A lot of works have exploited image/feature matching for many computer vision tasks, including optical flow estimation, stereo and correspondence matching. Due to the illumination robustness of Census Transform~(CT), it has been widely used in those research fields. For example, Stein \etal~\cite{stein2004efficient} use the Census Transform to convert the image patches from the RGB space to the CT-based feature space and verify its robustness of image matching on real-world cases. M{\"u}ller \etal~\cite{muller2011illumination} propose an illumination-invariant census transform for optical flow estimation. Hermann \etal ~\cite{hermann2013tv} adopt a census cost function for 3D medical image registration. In this work, we incorporate the robustness of census transform into our texture consistency loss to generate finer details.
	
	\subsection{Temporal Consistency}
	\zk{ Some previous studies~\cite{lai2018learning,zhang2020there,dwibedi2019temporal} have exploited temporal consistency for deep-learning models.  Lai~\etal~\cite{lai2018learning} present a short- and long-term temporal loss to enforce the model to learn consistent results over time. Recently, Zhang~\etal~\cite{zhang2020there} conduct extensive experiments to study spatial-temporal tradeoff for video super-resolution. Dwibedi~\etal~\cite{dwibedi2019temporal} utilize a temporal cycle-consistency loss for self-supervised representation learning.  In this work, we propose a novel temporal supervision which improves the quality of frame interpolation by considering adjacent frames. Following~\cite{zhang2020there}, we manually search for a balancing factor to achieve appropriate spatial-temporal tradeoff.
	}
	

	\section{The Proposed Video Interpolation Scheme}
	
	In this section, we first give an overview of the proposed algorithm for video frame interpolation (VFI) in Sec.~\ref{overview}. In Sec.~\ref{Matching}, we explain our texture consistency loss for supervision. Then, we elaborate on the cross-scale pyramid alignment and adaptive fusion in Sec.~\ref{align}. At last, we describe the configurations of our network in detail in Sec~\ref{network}.
	
	\vspace{-8pt}
	\subsection{Overview}
	\label{overview}
	Frame interpolation aims at synthesizing an intermediate frame (e.g., $I_{0}$) in the middle of two adjacent frames (e.g., $I_{-1}$ and $I_1$). As illustrated in Fig.~\ref{fig:framework}, our framework completes the interpolation in a four-step process. First, we obtain the feature pyramids $F_{-1}^{\{0,1,2\}}$ and $F_{1}^{\{0,1,2\}}$ of frames $I_{-1}$ and $I_1$ using a feature extraction module. After that, the extracted features are passed through a cross-scale pyramid alignment module to perform a bidirectional alignment towards the middle point in time. Then, we develop an attention-based fusion module to fuse aligned features $F_{-1 \rightarrow 0}$ and $F_{1 \rightarrow 0}$, resulting in $F_{0}$. Finally, a sequence of residual blocks are applied on $F_{0}$ to synthesize the intermediate frame $\hat{I}_{0}$.
	
	The existing methods usually strongly penalize the predicted frame $\hat{I}_{0}$ when it does not exactly match the predefined ground truth (GT) ${I}_{0}$. However, due to the non-uniqueness of movement between ${I}_{-1}$ and ${I}_{1}$, there may exist many plausible solutions in terms of ${I}_{0}$. Relaxing the rigid requirement of synthesizing the intermediate frame as close as possible to GT ${I}_{0}$, we allow the prediction to be supervised by not only the GT but also the corresponding patterns in ${I}_{-1}$ and ${I}_{1}$. In this case, our learning target is formulated as
	
	\begin{equation}
		{\hat I}_{0} =  \mathop{\arg\min}_{{{\hat I}_{0}}} ( \  L_1({{\hat I}_{0},{I}_{0}}) + \alpha L_{p}({{\hat I}_{0},{{I}_{-1},I_1}})),
		\label{opt0}
	\end{equation}
	where $L_1({{\hat I}_{0},{I}_{0}})$ is the commonly adopted data term and $L_{p}({{\hat I}_{0},{{I}_{-1},I_1}})$ is the proposed texture consistency loss detailed in Sec.~\ref{Matching}. The scaling parameter $\alpha$ is to balance the importance of the two items.
	
	\begin{figure}[t]
		\centering
		\includegraphics[width=1.0\columnwidth]{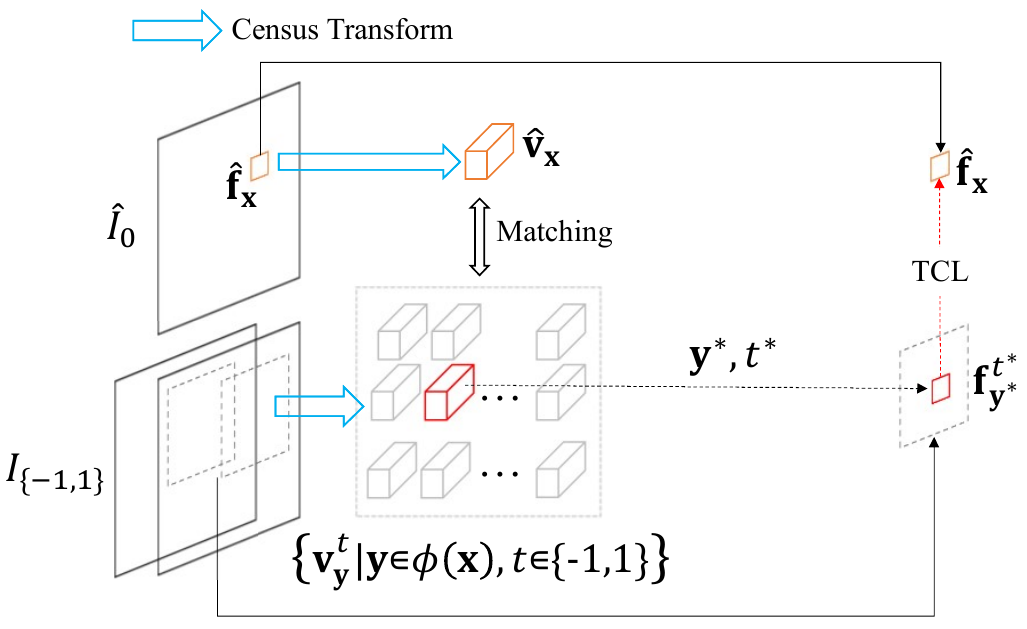} %
		
		\caption{Overview of our texture consistency loss~(TCL). The best matched $\mathbf{f}_{\mathbf{y}^*}^{t^*}$ is served as a pseudo label for training. } 
		\label{fig:matching}
		\vspace{-8pt}
	\end{figure} %
	\subsection{Texture Consistency Loss}
	\label{Matching}
	
	The proposed texture consistency loss is illustrated in Fig.~\ref{fig:matching}. For the patch $\mathbf{\hat f}_ {\mathbf{x}}$ centrally located at position $\mathbf{x}$ on the predicted frame ${\hat I}_{0}$, we first seek for its best matching $\mathbf{f}_{\mathbf{y}^*}^{t^*}$ from the input frames $\{I_{-1},I_1\}$, where $\mathbf{y}^*$ and $t^*$ are obtained from 
	\begin{equation}
		\mathbf{y}^*,t^* =  \mathop{\arg\min}_{\mathbf{y},t} \   L2(\mathbf{\hat f}_{\mathbf{x}},\mathbf{f}_{\mathbf{y}}^t),
	\end{equation}
	where $\mathbf{y}^*$ and $t^* \in \{-1,1\}$ refer to the optimal position and the optimal frame index, respectively. Then $\mathbf{f}_{\mathbf{y}^*}^{t^*}$ is adopted as an additional pseudo label for estimation $\mathbf{\hat f}_ {\mathbf{x}}$.
	
	In our implementation, to avoid the interference of illumination in the RGB space across frames, we first apply a census transform~\cite{zabih1996non} to the query and the matching candidates before matching:
	\begin{equation}
		\mathbf{v}_{\mathbf{x}}(\mathbf{x} + \mathbf{x}_n) =
		\begin{cases}
			0, & \mathbf{f}_{\mathbf{x}}(\mathbf{x}) > \mathbf{f}_{\mathbf{x}}(\mathbf{x} + \mathbf{x}_n)\\
			1, & \mathbf{f}_{\mathbf{x}}(\mathbf{x}) \le \mathbf{f}_{\mathbf{x}}(\mathbf{x} + \mathbf{x}_n) \\
		\end{cases},
		\mathbf{x}_n \in \mathcal{R},
	\end{equation}
	where $\mathbf{f}_{\mathbf{x}}(\mathbf{x})$ is the pixel value at centeral position $\mathbf{x}$ and the patch field $\mathbf{x}_n$ is defined as
	\begin{equation}
		\mathcal{R} = \{(-1, -1), (-1, 0), \dots, (0, 1), (1, 1)\}.
	\end{equation}
	To accelerate the matching process and maintain a reasonable receptive field, we further define the searching area as
	\begin{equation}
		{\mathop \phi}(\mathbf{x})=\left\{\mathbf{y} | \left| \mathbf{y}-\mathbf{x} \right| \le \mathbf{d} \right\},
	\end{equation}
	where $\mathbf{y}$ is the position of patch candidates and $\mathbf{d}$ indicates the maximum displacement. In this case, the matching process is represented as
	\begin{equation}
	\mathbf{y}^*,t^* = \mathop{\arg\min}_{ {\mathbf{y}}\in {\mathop \phi}(\mathbf{x}),t\in\{-1,1\} } \ L2(\mathbf{\hat v}_{\mathbf{x}},\mathbf{v}_{\mathbf{y}}^t), 
	\end{equation}
	where $\mathbf{\hat v}_\mathbf{x}$ and $\mathbf{v}_\mathbf{y}^t$ are the representations of patches $\mathbf{\hat f}_\mathbf{x}$ and $\mathbf{f}_\mathbf{y}^t$ after census transform. Noticing that the operation of census transform is non-differentiable, our TCL is performed on the original RGB space as
	\begin{equation}
		L_{p}({{\hat I}_{0},{I}_{-1},I_1})(\mathbf{x}) = L1(\mathbf{\hat f}_\mathbf{x},\mathbf{f}_{\mathbf{y}^*}^{t^*}).
	\end{equation}


	\subsection{Cross-Scale Pyramid Alignment}
	\label{align}

	As aforementioned in Sec.~\ref{sec:intro}, most VFI methods utilize the optical flow to perform a two-step synthesis, image-level alignment and deep-learning-based interpolation. However, these approaches face challenges in handling occluded or textureless areas. Consequently, the inaccurate alignment may degrade the performance of the latter processing phases. By contrast, kernel-based works formulate the interpolation as an adaptive convolution over input frames, which typically use a deep network to regress a pair of pixel-wise kernels and apply them on the input frames. However, this single-scale aggregation at the image level may not make full use of information of input frames. To cope with this problem, some approaches have exploited multi-scale aggregation strategies by building dense correlation maps. Nevertheless, the computational complexity increases dramatically with the growth of image resolution. 
	
	In this work, we develop a cross-scale pyramid alignment~(CSPA) at the feature level aided by deformable convolution networks~\cite{dai2017deformable,zhu2019deformable}. Compared with the previous multi-scale aggregation strategies, CSPA has the following advantages: (1) the previous aligned low-resolution results are regarded as a guidance for the alignment of higher-resolution features, which ensures more accurate warping; (2) aggregating cross-scale information is beneficial to restoring more details; (3) without constructing a cost volume or correlation map, our CSPA is more computationally efficient. 
			
	\begin{figure*}[t]
		\centering
		\includegraphics[width=0.95\linewidth]{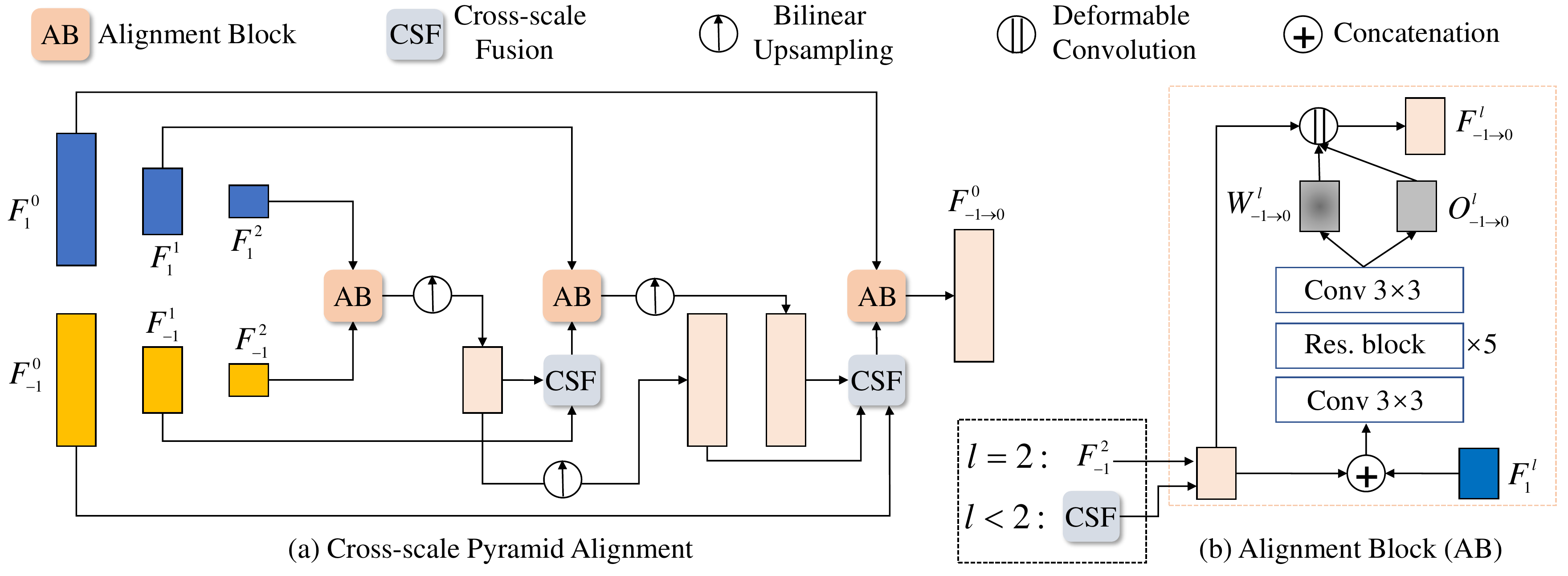} 
		\caption{The framework of the cross-scale pyramid alignment~(CSPA) module. (a) The $3$-level pyramid alignment from $I_{-1}$ to $I_{0}$. (b) The detailed structure of the alignment block at the $l$-th level. The source feature~(in light yellow) and the other endpoint feature $F_{1}^{l}$ (in blue) are fed through the alignment block to generate an aligned feature $F_{-1\rightarrow 0}^l$. The source feature is $F_{-1}^2$ when $l=2$, while representing the fused result of CSF for other cases. More details can be found in Section~\ref{align}.}
		\vspace{-0.15in}
		\label{fig:cspa}
	\end{figure*} %
	
	In detail, the feature pyramids $F_{-1}^{\{0,1,2\}}$ and $F_1^{\{0,1,2\}}$ of frames $I_{-1}$ and $I_1$ are aligned in a bidirectional way. Taking the direction of $I_{-1} \rightarrow I_{0}$ for example, we gradually align $F_{-1}^{\{0,1,2\}}$ from low resolution (i.e., $F_{-1}^2$) to high resolution (i.e., $F_{-1}^0$), as illustrated in Fig.~\ref{fig:cspa}. At first, referring to the other endpoint $F_{1}^2$, the alignment of $F_{-1}^2$is conducted to handle the large motion as
	\begin{equation}
		F_{-1 \rightarrow 0}^2 = Align(F_{-1}^2,F_1^2).
		\label{eq:align}
	\end{equation}
	Next, at the higher resolution level, we bilinearly upsample $F_{-1 \rightarrow 0}^2$ to ${F}_{-1 \rightarrow 0}^{2,\uparrow 2}$ by a factor of 2 and aggregate the cross-scale (level 2 and level 1) information as
	\begin{equation}
		\tilde F_{-1 \rightarrow 0}^1 = {Fuse}({F}_{-1 \rightarrow 0}^{2,\uparrow 2}, F_{-1}^{1}).
		\label{eq:f1}
	\end{equation}
	in a fusion module (``CSF'' in Fig.~\ref{fig:cspa}a), which is implemented as a concatenation followed by a convolution operation.
	Later on, we feed the fusion $\tilde F_{-1 \rightarrow 0}^1$ and $F_1^1$ at the other endpoint into the alignment block and obtain the aligned result:
	\begin{equation}
		F_{-1 \rightarrow 0}^1 = Align(\tilde F_{-1 \rightarrow 0}^1, F_1^1).
		\label{eq:align}
	\end{equation}
	Following the same pipeline, we perform the alignment at the highest resolution level to handle subtle motion. Specifically, the cross-scale fusion module takes three-level inputs as 
	\begin{equation}
		\tilde F_{-1 \rightarrow 0}^0 = Fuse({F}_{-1 \rightarrow 0}^{2,\uparrow 4}, {F}_{-1 \rightarrow 0}^{1,\uparrow 2}, F_{-1}^{0}).
		\label{eq:f1}
	\end{equation}
	It is noted that the alignment of  $I_{1} \rightarrow I_{0}$ is completed symmetrically.

	The alignment block is zoomed up in Fig.~\ref{fig:cspa}b. In terms of the $l$-th level alignemnt for frame $I_{-1}$, the block first concatenates the fused cross-scale feature $\tilde F_{-1 \rightarrow 0}^l$ and $F_1^l$, and conducts a $3 \times 3$ convolution. Five sequential residual blocks and another convolution are used to predict a weight map $W_{-1 \rightarrow 0}^l$ and an offset map $O_{-1 \rightarrow 0}^l$. Finally, the aligned feature $F_{-1 \rightarrow 0}^l(\mathbf{x})$ at position $\mathbf{x}$ is calculated by
	\begin{equation}
		F_{-1 \rightarrow 0}^l(\mathbf{x}) = \sum_{i} \tilde F_{-1 \rightarrow 0}^l(\mathbf{x} + O_{-1 \rightarrow 0, i}^l(\mathbf{x})) * W_{-1 \rightarrow 0, i}^l(\mathbf{x}),
		\label{dcn}
	\end{equation}
	where the subscript $i$ means the $i$-th element in the receptive field of convolution.

\begin{figure}[h]
		\centering
		\includegraphics[width=1.0\columnwidth]{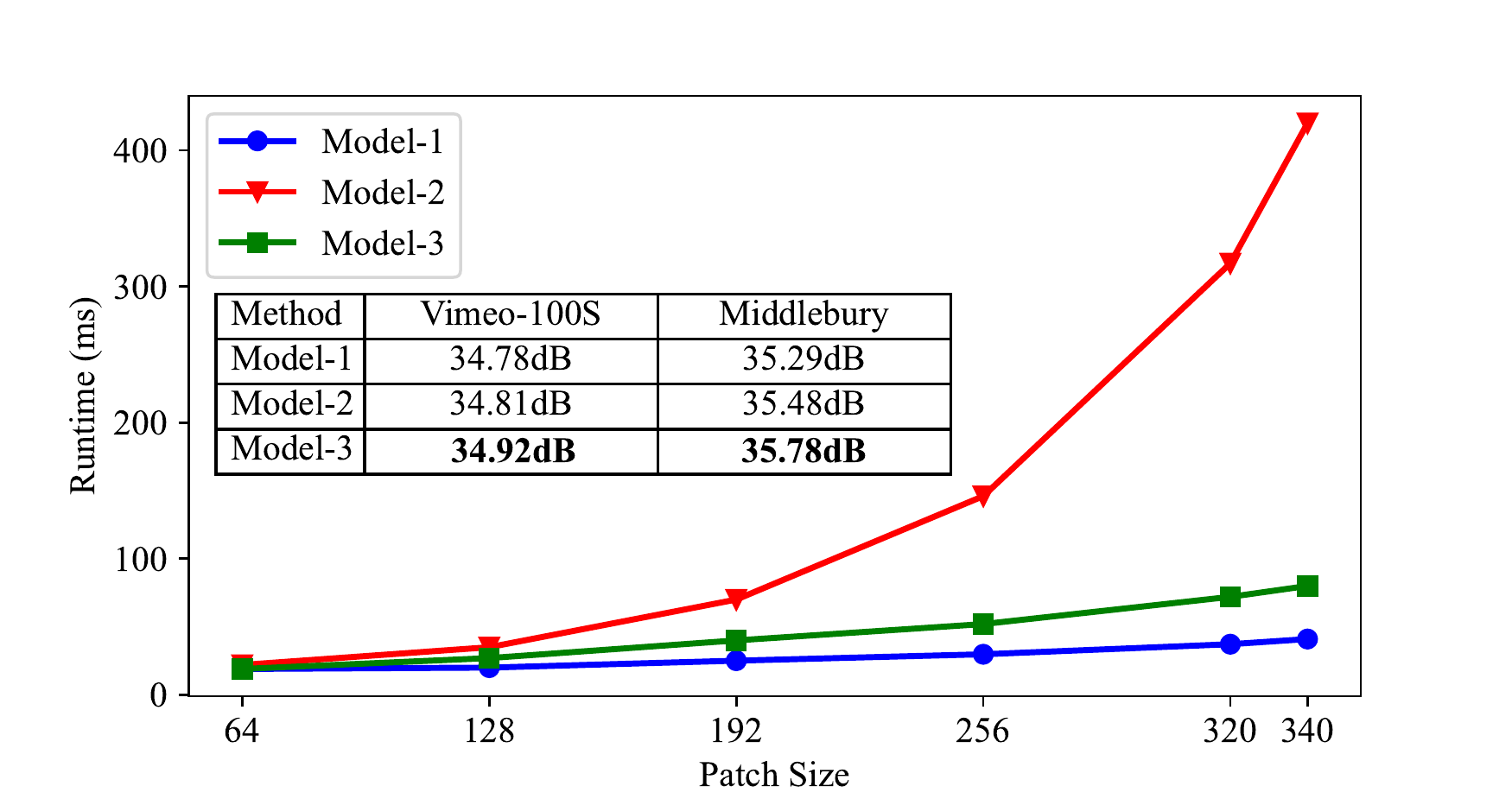} 
		\caption{Efficiency of three alignment models. Model-1 refers to the single-scale alignment. Model-2 denotes the cross-scale alignment using a cost volume. Model-3 is our proposed cross-scale pyramid alignment~(CSPA). Our CSPA achieves the most promising results, while only requiring comparable computational complexity with the single-scale alignment.}
		\label{fig:rt_cspa}
		\vspace{-0.15in}
	\end{figure}

	As detailed in the Appendix~\ref{sec:AppenB}, the proposed CSPA module is of $O(N)$ computational complexity where $N$ is the number of pixels. In Fig.~\ref{fig:rt_cspa}, we compare the running time (on NVIDIA RTX 2080Ti) of three alignment models: single-scale~(Model-1), cross-scale using a cost volume~(Model-2) and the proposed CSPA~(Model-3). The complexities of Model-1 and Model-2 are $O(N)$ and $O(N^2)$. It is observed that our CSPA costs much less time than the cost-volume-based Model-2 for the large-scale input. Especially, CSPA obtains comparable efficiency compared to the single-scale Model-1. In terms of the GPU memory cost for $340 \times 340$ input, our model only requires 4.0 GBytes that is close to the Model-1~(2.3 GBytes), much smaller than the Model-2~(10.0 GBytes). To quantitatively evaluate the performance of three strategies, we train three models under a fair experimental setting\footnote{For a quick comparison, we construct a small training set that contains 5000 samples from Vimeo-Triplets~\cite{xue2019video} and randomly select 100 testing samples from Vimeo-Triplets-Test~(denoted as ``Vimeo-100S"). We also test the generalization ability of three models by assessing their performance on the out-of-domain Middlebury set~\cite{baker2011database}.}. The PSNR results shown in Fig.~\ref{fig:rt_cspa} further demonstrate the effectiveness of our CSPA approach~(Model-3).

	\subsection{Attention-Based Fusion}
	\label{fusion}
	After the bidirectional alignment, we obtain a pair of aligned features $F_{-1 \rightarrow 0}^0$ and $F_{1 \rightarrow 0}^0$. In order to determine whether the information is useful or not in a spatially variant way, we employ an attention mechanism to aggregate these two features. First, the attention map is calculated by a convolution followed by a sigmoid operation as
	\begin{equation}
		M = Sigmoid( Conv(F_{-1 \rightarrow 0}^0,F_{1 \rightarrow 0}^0)).
	\end{equation}
	Then, the final aggregated result $F_{0}$ is obtained by
	\begin{equation}
		F_{0} = M*F_{-1 \rightarrow 0}^0 + (1-M)*F_{1 \rightarrow 0}^0.
	\end{equation}

	\subsection{Other Configurations}
	\label{network}
	\paragraph{Common Settings}
	As illustrated in Fig.~\ref{fig:framework}, there are four modules in our framework: feature extraction, cross-scale pyramid alignment, attention-based feature fusion, and reconstruction. The number of parameters are 4.28M, 12.52M, 0.29M and 11.80M, respectively, in a total of 28.89M. Following~\cite{wang2019edvr}, we adopt the residual block~\cite{he2016deep} as the basic component~(shorted as ``RB"), which is detailed in Table~\ref{tab:residual}. In our network, the channel number of convolutions is set to 128. We use $\Rightarrow$ to point out the output of a layer in~\cref{tab:residual,tab:feaext,tab:recons}.
	\begin{table}[h]	
		\begin{center}
			\begin{tabular}{|c|c|c|c}
				\hline
				Input        	& x \\ \hline
				Layer1       	& Conv(128,128,3,1) + ReLU \\ \hline
				Layer2       	& Conv(128,128,3,1) $\Rightarrow y$\\ \hline 
				Output       	& x+y   \\ \hline 
				Params.   & 0.3M   \\ \hline 
			\end{tabular}
		\end{center}
		\caption{The structure of the residual block~(``RB"). }
		\label{tab:residual}
		\vspace{-8pt}
	\end{table}
	
	\paragraph{Feature Extraction}
	The structure of the feature extraction module is shown in Table~\ref{tab:feaext}. For a given input frame $I_i \in R^{ C \times H\times W}$~($i=\{-1,1\}$), we first utilize a convolution to change its channel dimension to 128. Then the feature maps are passed through five residual blocks, resulting in the $0$-th level feature $F_i^0$ of the pyramid representation. Finally, we use two convolutional layers with strides of 2 to generate the downsampled features $F_i^{1}$ and $F_i^{2}$, respectively.
	\begin{table}[h]	
		\begin{center}
			\begin{tabular}{|c|c|c}
				\hline
				Input        & $I_i$  \\ \hline
				Layer1       & Conv(3,128,3,1) + ReLU \\ \hline
				Layer2       & $5\times$ RB(128) $\Rightarrow F_i^{0}$\\ \hline 
				Layer3       & Conv(3,128,3,2) + ReLU $\Rightarrow F_i^{1}$\\ \hline 
				Layer4       & Conv(3,128,3,2) + ReLU  $\Rightarrow F_i^{2}$\\ \hline 
				Params.   & 4.28M   \\ \hline 
			\end{tabular}
		\end{center}
		\caption{The structure of our feature extraction module.}
		\label{tab:feaext}
	\end{table}
	
	\paragraph{Reconstruction}
	Table~\ref{tab:recons} shows the details of the reconstruction module. The fused intermediate feature $F_{0}$ is firstly passed to a sequence of residual blocks for refinement. At last, we use a single convolution without activation to generate the final result $I_{0}$.
	\begin{table}[h]	
		\begin{center}
			\begin{tabular}{|c|c|}
				\hline
				Input        	& $F_{0}$  \\ \hline
				Layer1       	& $40\times$RB(128) \\ \hline
				Layer2       	& Conv(128,3,3,1) $\Rightarrow {\hat I}_{0}$\\ \hline 
				Params.   &11.80M   \\ \hline 
			\end{tabular}
		\end{center}
		\caption{The structure of the reconstruction module.}
		\label{tab:recons}
	\end{table}
	\begin{figure}[h]
		\centering
		\includegraphics[width=1.0\columnwidth]{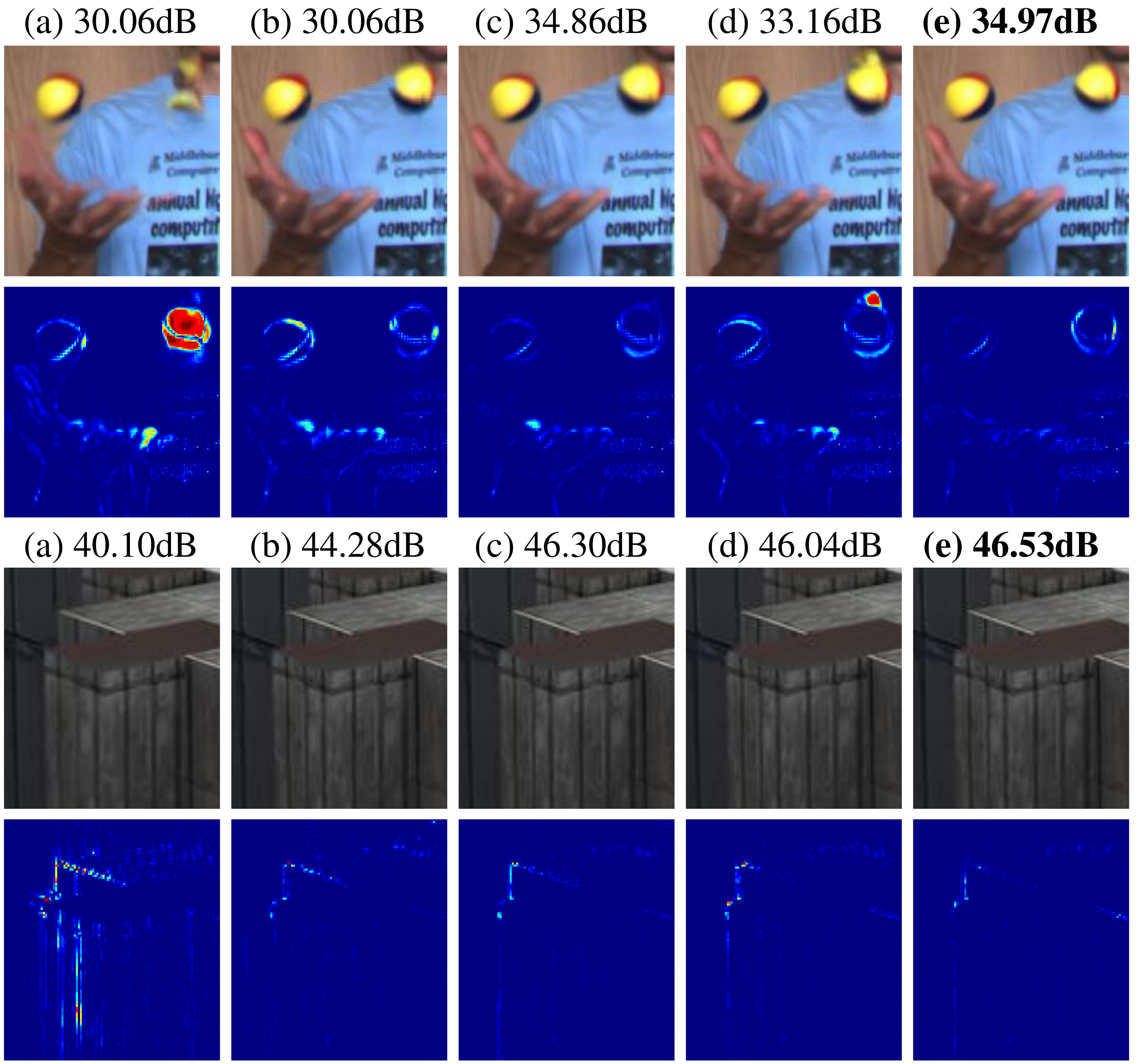} 
		
		\caption{Visualized results for single-frame interpolation on Middlebury~\cite{baker2011database}. (a-e) represents SepConv~\cite{niklaus2017videosep}, CtxSyn~\cite{niklaus2018context}, Softmax-Splatting~\cite{niklaus2020softmax}, RIFE-L~\cite{huang2020rife} and ours. The first and third rows show some cropped regions of the interpolated images, while the second and fourth rows are the corresponding error maps normalized between $[0,1]$ for best view. The best results are highlighted in {\bf bold}.} 
		\vspace{-0.15in}
		\label{fig:tri_sota_mid}
	\end{figure} %
	\begin{figure*}[h]
		\centering
		\includegraphics[width=1.0\linewidth]{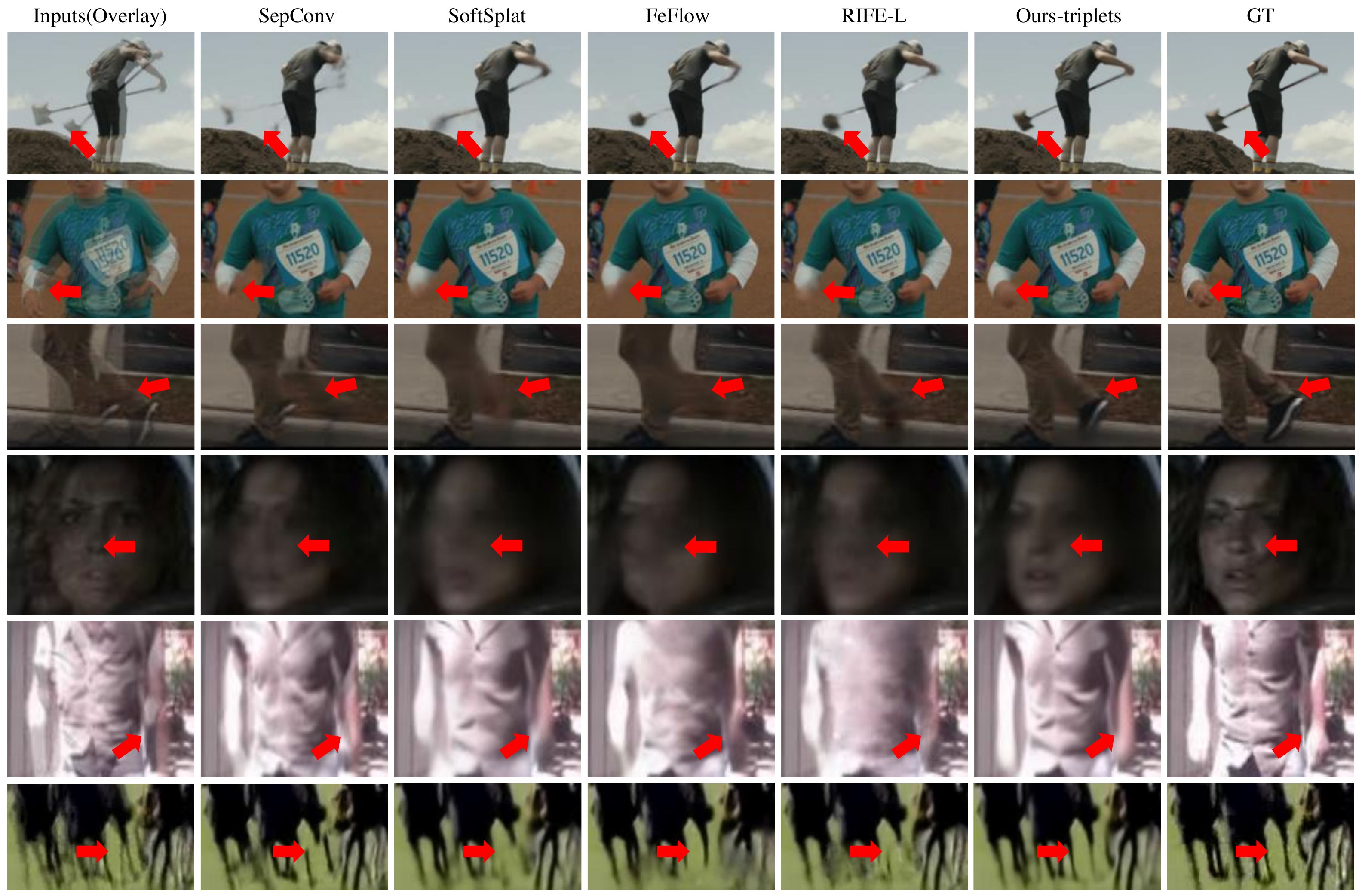} 
		
		\caption{Visual comparison of single-frame VFI algorithms. Our method outperforms other state-of-the-art approaches with finer details and fewer artifacts.} 
		\label{fig:tri_sota}
		\vspace{-0.15in}
	\end{figure*} %
	
	\begin{table*}[t]	
		\small
		\setlength{\tabcolsep}{6pt}
		\begin{center}
			\begin{tabular}{cccccccccc}
				\hline
				\multirow{2}{*}{Method} & training  & \# Parameters & Runtime  &\multicolumn{2}{c}{Vimeo-Triplets-Test}    &\multicolumn{2}{c}{Middlebury} &\multicolumn{2}{c}{UCF101} \\ \cmidrule(r){5-6} \cmidrule(r){7-8} \cmidrule(r){9-10}
				&dataset    &  (Million)   &  (ms)   & PSNR     &SSIM               & PSNR     &SSIM       & PSNR     &SSIM 		\\ \hline
				TOFlow~\cite{xue2019video}         &Vimeo-Triplets-Train       &1.1 &72               &33.73     &0.968    &35.29     &0.956        & 34.58    &0.967                  \\ 
				SepConv~\cite{niklaus2017videosep} &proprietary                &21.6 &51            &33.79     &0.970     &35.73      &0.959             & 34.78    &0.967     		\\ 
				SoftSplat~\cite{niklaus2020softmax}  &Vimeo-Triplets-Train     &7.7 &135             &{\color{blue} 36.10}     &{\color{blue} 0.980}  &{\color{blue}38.42}     &0.971               & {\color{blue}35.39}    &{\color{blue}0.970}           \\ 	
				BMBC~\cite{park2020bmbc}	 &Vimeo-Triplets-Train             &11.0 &1580             &35.01     &0.976     &36.66   &0.983                & 35.15    &0.969                  		 \\ 
				DSepConv~\cite{cheng2020video}	& Vimeo-Triplets-Train         &21.8 &236              &34.73     &0.974       & -  & -              & 35.08    &0.969               		  \\ 
				DAIN~\cite{bao2019depth}        &Vimeo-Triplets-Train          &24.0 &130            &34.71     &0.976   &36.70     &0.965               & 35.00    &0.968              \\ 
				CAIN~\cite{choi2020channel}     &Vimeo-Triplets-Train          &42.8 &38             &34.65     &0.973      & -  & -           & 34.98    &0.969                       		 \\ 
				EDSC~\cite{cheng2021multiple}	&Vimeo-Triplets-Train          &8.9 &46               &34.84     &0.975      &36.80      &0.983          & 35.13    &0.968             \\ 
				$\dagger$SepConv++~\cite{niklaus2021revisiting}                &Vimeo-Triplets-Train         &-  & -                 &34.98    & -         &37.47      &-             &35.29      &-             \\
				PWDN~\cite{chen2021pdwn}         &Vimeo-Triplets-Train         &7.8  & -                &35.44    & -         &37.20     &-             &35.00      &-             \\
				FeFlow~\cite{gui2020featureflow}  &Vimeo-Triplets-Train         &133.6  & -                &35.28    & -         &-     &-             &35.08     &0.957             \\
				MEMC-Net~\cite{bao2019memc}       &Vimeo-Triplets-Train        &70.3 &120              &34.40     &0.970      & -  & -           & 35.01    &0.968                      		 \\ 
				RIFE-L~\cite{huang2020rife}      &Vimeo-Triplets-Train         &20.9 &72               &{\color{blue} 36.10}     &{\color{blue} 0.980}   &37.64  &{\color{blue}0.985}               & 35.29    &0.969                   		  \\ \hline
				
				Ours-triplets                             &Vimeo-Triplets-Train         &28.9 &680        &{\color{red} 36.76}     &{\color{blue}0.980}             &{\color{red}38.83} &    {\color{red}0.989}                &{\color{red}35.43}   &{\color{red}0.979}          \\ \hline
			\end{tabular}
		\end{center}
		\caption{Quantitative comparison of single-frame VFI algorithms. The numbers in red and blue refer to the best and second-best PSNR(dB)/SSIM results. Runtime of each model is also reported with an input size of $2\times480\times640$. ``$\dagger$" means using self-ensembling during evaluation. }
		\vspace{-0.1in}
		\label{tab:tri_sota}
	\end{table*}
	
	\begin{figure}[h]
		\centering
		\includegraphics[width=1.0\columnwidth]{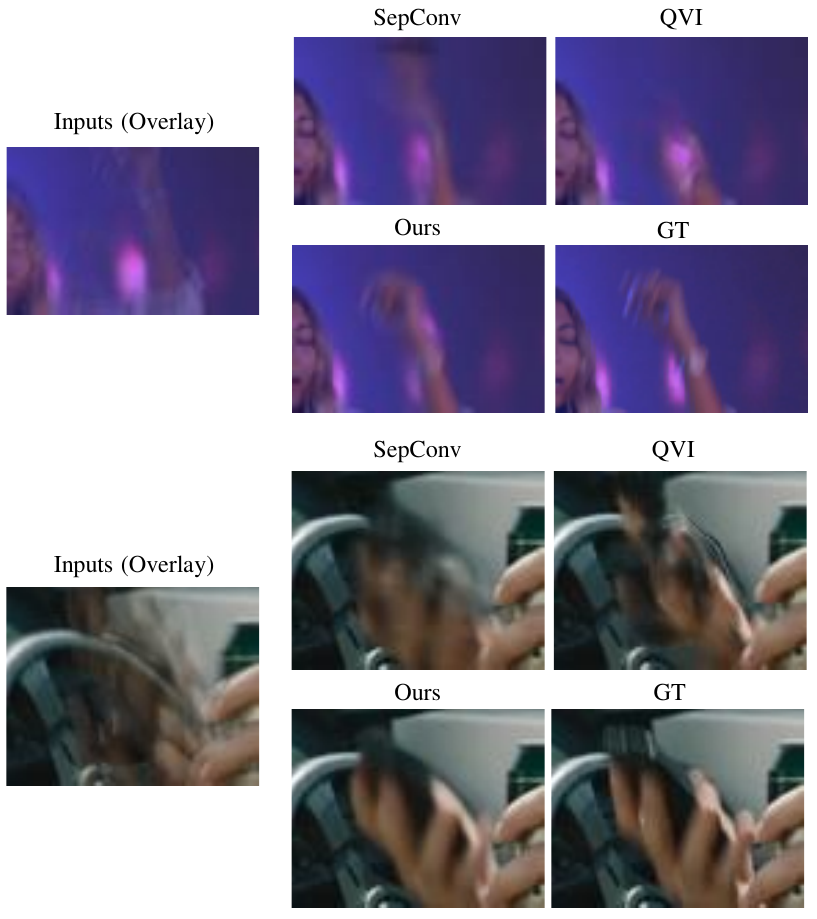} %
		
		\caption{Visual comparison of multi-frame VFI algorithms. } 
		\vspace{-0.25in}
		\label{fig:mfi}
	\end{figure} %
	\vspace{-0.15in}
	\section{Experiments}
	\subsection{Implementation Details}
	All experiments are conducted on the NVIDIA GeForce RTX 2080Ti GPUs. We use two adjacent frames to interpolate the middle frame. An Adam optimizer is adopted and the learning rate decays from $5 \times 10^{-4}$ to $0$ by a cosine annealing strategy. We set the batch size to 64. The training lasts 600K iterations, during which we adopt random $64 \times 64$ cropping, vertical or horizontal flipping, and $90^{\circ}$ rotation augmentations.

	\subsection{Datasets and Evaluation Metrics}
	\label{data}
	\paragraph{Vimeo-Triplets~\cite{xue2019video}}
	It contains 51,312 and 3,782 triplet frames with a resolution of $256 \times 448$ for training and testing, respectively. Following the most commonly used protocols~\cite{xue2019video,niklaus2020softmax,park2020bmbc,cheng2020video,bao2019depth,choi2020channel,cheng2021multiple,bao2019memc,huang2020rife}, we train a model for single-frame interpolation on the training split~(Vimeo-Triplets-Train) while evaluating the results on the testing part~(Vimeo-Triplets-Test).

	\paragraph{Middlebury~\cite{baker2011database}}
	It is a widely used evaluation benchmark for video frame interpolation. In total, there are 12 challenging cases where each of them contains three video frames. The central frame serves as the ground truth while the others are used as the input.
	
	\paragraph{UCF101~\cite{soomro2012dataset}}
	It is a large scale dataset for human action recognition, where Liu \etal~\cite{liu2017video} sample 379 triplets ($256 \times 256$) for single-frame interpolation evaluation.  Unlike the aforementioned datasets, UCF101 has heavy compression noises. Following recent methods~\cite{xue2019video,niklaus2020softmax,park2020bmbc,cheng2020video,bao2019depth,choi2020channel,cheng2021multiple,bao2019memc,huang2020rife}, we assess our approach on this benchmark without finetuning.
	
	\paragraph{UCF101-E~\cite{NEURIPS2019_d045c59a}}
	Since UCF101 can only be used for evaluating single-frame interpolation, QVI~\cite{NEURIPS2019_d045c59a} further collects the UCF101-E benchmark for multi-frame interpolation. UCF101-E is composed of 100 samples and each sample has 5 consecutive frames with a resolution of $225\times225$.
	
	\paragraph{ Vimeo90K-7f~\cite{xue2019video}}
	It is widely used in the video super-resolution task. There are 64,612 training and 7,824 validation video sequences containing 7 consecutive frames with a resolution of $256\times 448$. In this work, we train our multi-frame interpolation model using the training split of  Vimeo90K-7f. In detail, we randomly sample 5 consecutive frames at a time where the central frame is the interpolation target and the others are used as the input.
	
	\paragraph{Metrics}
	We adopt PSNR and SSIM~\cite{wang2004image} as the quantitative evaluation metrics. The higher values indicate the better results.
	
	\vspace{-0.1in}
	\subsection{Comparison with SOTA Methods}
	
	To verify the effectiveness of the proposed method, we make a comparison with state-of-the-art methods under single-frame and multi-frame interpolation settings.
	
	\paragraph{Single-Frame Interpolation}
	As illustrated in Table~\ref{tab:tri_sota}, it is clear that our model achieves a new state of the art on all benchmarks. While some methods may incorporate additional information (e.g., optical flow, depth), our method still stands out as the best, especially outperforming the second-best SoftSplat~\cite{niklaus2020softmax} by 0.66dB and 0.41dB on Vimeo-Triplet-Test and Middlebury. For example, SoftSplat~\cite{niklaus2020softmax} relies on accurate bidirectional optical flow, RIFE-L~\cite{huang2020rife} requires a optical flow supervision, and DAIN~\cite{bao2019depth} utilizes additional depth information. Meanwhile, due to the heavy compression nature of images in UCF101, our model only obtains a slight improvement on PSNR/SSIM, indicating that the domain gap is a critical issue in the frame interpolation task. 
	
	
	We also show some visual examples in Fig.~\ref{fig:tri_sota} and Fig.~\ref{fig:tri_sota_mid}. Compared with other methods, our model successfully handles complicated motion and produces more plausible structures. In terms of the second example in Fig.~\ref{fig:tri_sota}, all other methods fail to restore the right hand of the human, while ours interpolates the hand that is closest to the ground truth. As for the first example in Fig.~\ref{fig:tri_sota_mid}, both SepConv~\cite{niklaus2017videosep} and RIFE-L~\cite{huang2020rife} cannot synthesize regular balls. Although Context-Syn~\cite{niklaus2018context} and SoftSplat~\cite{niklaus2020softmax} generate visually correct balls, they produce some annoying artifacts on the hand. On the contrary, our method can interpolate the contents well without generating notable artifacts. The error maps further demonstrate the effectiveness of our method.
	
	\paragraph{Multi-Frame Interpolation}
	\begin{table}[t]	
		\small
		\setlength{\tabcolsep}{3pt}
		
		\begin{center}
			\begin{tabular}{cccccccccc}
				\hline
				\multirow{2}{*}{Method} & \# Parameters & \multicolumn{2}{c}{Vimeo90K-7f}    &\multicolumn{2}{c}{UCF101-E} \\ \cmidrule(r){3-4} \cmidrule(r){5-6}
				&  (Million)      & PSNR     &SSIM               & PSNR     &SSIM      		\\ \hline
				AdaCof~\cite{lee2020adacof}               &21.8                 &33.92     &0.945              & -    &-          \\ 
				SepConv~\cite{niklaus2017videosep}       &21.6               &33.65     &0.943              & 31.94   &0.942     		\\ 
				
				DVF~\cite{liu2017video}                  &3.8                &30.79     &0.891              & 29.88    &0.916           \\ 
				Slomo~\cite{jiang2018super}            &39.6                 &33.73     &0.945              & -    &-                   		 \\ 
				Phase~\cite{meyer2015phase}       &-               &30.52     &0.885              &-    &-          \\ 	
				
				QVI~\cite{NEURIPS2019_d045c59a}              &29.2                 &35.19    &0.956              & 32.54    &0.948                    		  \\ \hline
				Ours-triplets                &28.9 &{\color{blue}36.13}  &{\color{blue}0.960}  &{\color{blue}32.91} &{\color{red}0.952}\\
				Ours                                    &28.9                 &{\color{red} 36.50}     &{\color{red}0.962}              &{\color{red}33.08}   &{\color{blue}0.951}            \\ \hline
			\end{tabular}
		\end{center}
		\caption{Quantitative comparison of multi-frame VFI algorithms. The numbers in red and blue refer to the best and second-best PSNR(dB)/SSIM results.}
		\label{tab:quad_sota}
	\end{table}

	To compare with methods that utilize multiple input frames on both sides, e.g., generating $I_{0}$ from $\{I_{-2},I_{-1},I_{1},I_{2}\}$, we follow~\cite{NEURIPS2019_d045c59a} to train a model on the Vimeo90k-7f training subset.
	
	In Table~\ref{tab:quad_sota}, we quantitatively evaluate our method on the Vimeo90K-7f~\cite{xue2019video} and UCF101-E~\cite{soomro2012dataset} benchmarks. It is observed that the performance of our method surpasses all competing methods including kernel-based models~\cite{niklaus2017videosep,lee2020adacof}, optical flow-based models~\cite{liu2017video,jiang2018super} and motion model~\cite{NEURIPS2019_d045c59a}. Especially, our method brings about nearly 1.3dB PSNR improvement over the second-best QVI~\cite{NEURIPS2019_d045c59a} on the Vimeo90K-7f testing set. Besides, it is noteworthy that our model with only two input frames (denoted as``Ours-triplet" ) still outperforms other methods that take four frames as input.
	
	Fig.~\ref{fig:mfi} shows the qualitative comparison between our method and the other two approaches. For cases with large motion, the kernel-based SepConv~\cite{niklaus2017videosep} suffers from the restricted receptive field thus is difficult to handle the lone-range correspondences. As a result, blurry contents are sometimes produced. As for the flow-based method QVI~\cite{NEURIPS2019_d045c59a}, it may generate artifacts for some challenging cases. By contrast, our approach interpolates higher-quality frames with sharper edges, demonstrating the effectiveness of our proposed method.
	
	\subsection{Ablation Study}
	In this section, we make a comprehensive analysis of the contribution of each proposed component under the single-frame interpolation setting. The Vimeo-Triplets-Test is adopted as the evaluation benchmark.
	
	\begin{table}[t]
		\small
		\setlength{\tabcolsep}{6pt}

		\begin{center}
		
			\begin{tabular}{c|c|c } 
				\hline
				Method                      & PSNR~(dB) &SSIM    \\ \hline
				
				Baseline             			 &  35.90      		     &  0.969  \\ 
				Baseline {w/} TCL                 &  36.21(+0.31)         &   0.977(+0.08)  \\  
				Baseline {w/} CSPA                 &  36.56(+0.66)         &  0.976(+0.07)   \\
				Full                             &  36.76(+0.86)         &  0.980(+0.11)   \\  \hline
			\end{tabular}
		\end{center}
		\caption{Ablation studies of the proposed components on the Vimeo-Triplets-Test set.  }
		\label{table:ablation}
		\vspace{-0.15in}
	\end{table}
	%
	\begin{figure}[h]
		\centering
		\begin{subfigure}{\columnwidth}
			\includegraphics[width=1.0\columnwidth]{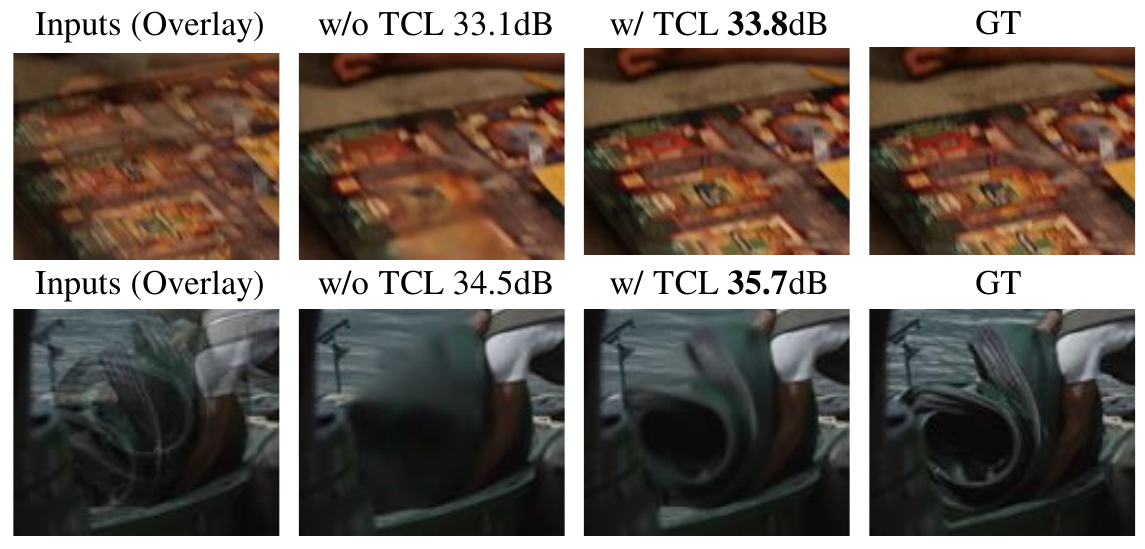}  %
			
			\caption{Visual comparison of results with/without TCL. } 
			\label{fig:oloss}
			
		\end{subfigure}
		\hfill
		\begin{subfigure}{\columnwidth}
			\includegraphics[width=1.0\columnwidth]{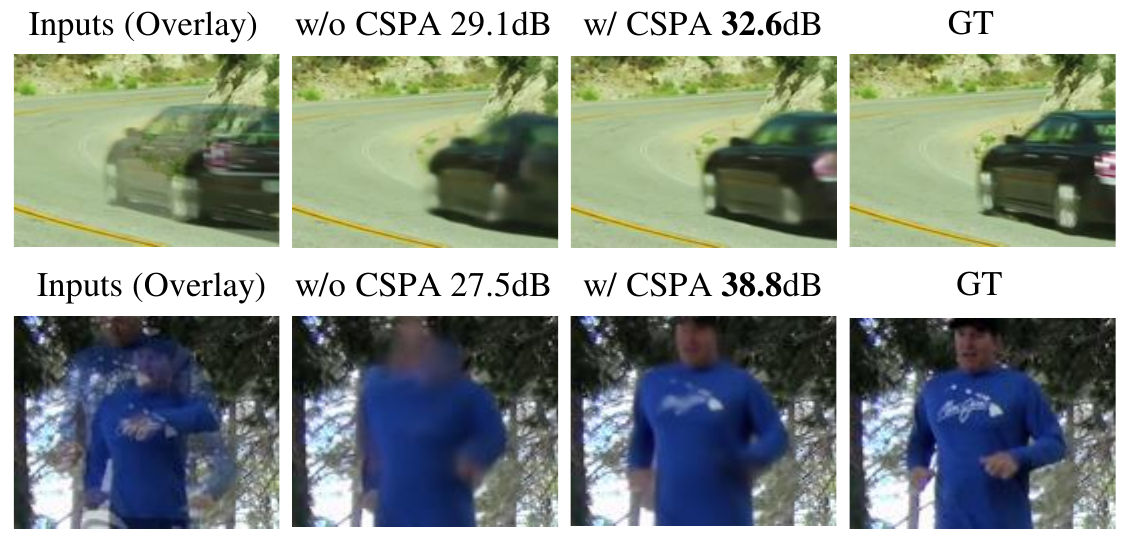}  %
			
			\caption{Visual comparison of results with/without CSPA.  } 
			\label{fig:abcspa}
		\end{subfigure}
		\caption{Ablation study of TCL and CSPA in visualization.}
		\vspace{-0.1in}
	\end{figure}


	\paragraph{Texture Consistency Loss (TCL)}
	The proposed TCL is designed to alleviate the over-constrained issue of the predefined ground truth, which actually is just one of many possible solutions given observed input frames. To verify this claim, we quantitatively compare the conventional L1 loss and the proposed TCL. As illustrated in Table~\ref{table:ablation}, the baseline trained with the additional TCL achieves a better performance in terms of PSNR and SSIM compared to the baseline. We also give some visual examples to illustrate the impact of the proposed TCL in Fig.~\ref{fig:oloss}. The model trained with TCL is able to preserve the structures of interpolated contents.
	
	\paragraph{Hyperparamter $\alpha$ in Eq.{\rm ~\ref{opt0}}}
	The hyperparameter $\alpha$ is used to balance the predefined ground truth and our proposed pseudo label. From Table ~\ref{table:alpha}, we notice that $\alpha=0.1$ is the best setting in our experiments (may not be optimal), and a large $\alpha$ harms the performance of TCL. Especially, the model trained with $\alpha=10.0$ fails to converge. We think of the proposed pseudo label better serving as auxiliary supervision apart from the L1 loss.
	
	%
	
	\begin{table}[h]
		\centering
		\small
		\begin{tabular}{c|c|c|c|c|c|c } 
			\hline
			$\alpha$        &0            & 0.1  &0.5        &1.0         &2.0 &10.0 \\ \hline
			
			PSNR~(dB)       &36.56      & {\bf 36.76}     &36.69   & 36.69            &  36.54       &- \\ 
			SSIM       &0.976   & {\bf 0.980}     & 0.979  & 0.979            & 0.978      &-    \\  \hline
		\end{tabular}
		\caption{  Analysis of $\alpha$ in  Eq.~\ref{opt0} on the Vimeo-Triplets-Test benchmark.  }
		\label{table:alpha}
	\end{table}
	
	\paragraph{Patch size $K$ of TCL}
	
	We explore the influence of the patch size $K \in {\{3,5,7,9\}}$ used in the TCL. As shown in Table~\ref{table:patchsize}, a larger patch size may degrade the performance of the model. It is reasonable since the increase in patch size brings more difficulties in matching correctly to the candidates on neighboring frames and the inaccurate supervision signals bring negative impacts during training.

	\begin{table}[h]
		\centering
		\small
		\setlength{\tabcolsep}{6pt}
		\begin{tabular}{c|c|c|c|c } 
			\hline
			$K$                    & 3          &5            &7          &9 \\ \hline
			
			PSNR~(dB)       &{\bf 36.76}          & 36.64              &  36.56     &  36.50  \\ 
			SSIM        &{\bf 0.980}           & 0.979             &  0.978     &  0.978  \\  \hline
		\end{tabular}
		\caption{Analysis of different patch sizes in TCL on the Vimeo-Triplets-Test benchmark.  }
		\label{table:patchsize}
	\end{table}
	
	\paragraph{Census Transform}
	We analyze the effect of adopting census transform in our TCL. We train another model by performing patch matching in the RGB space directly (denoted as ``TCL-RGB"). The results are described in Table~\ref{table:ct_rgb}. It is observed that ``TCL-RGB" leads to a lower interpolation quality in terms of PSNR and SSIM, which supports the claim that census transform is useful in eliminating the interference of illumination in Sec.~\ref{Matching}.
	\begin{table}[h]
		\centering
		\small
		\setlength{\tabcolsep}{9pt}
		\begin{tabular}{c|c|c} 
			\hline
			Method                  & Vimeo-Triplets-Test          &Middlebury           \\ \hline
			
			TCL-RGB       & 36.57/0.978 & 38.41/0.988  \\ 
			TCL-CT        & \bf{36.76/0.980} & \bf{38.83/0.989}             \\  \hline
		\end{tabular}
		\caption{ Analysis of cencus transform in TCL. We adopt PSNR(dB) and SSIM as the evaluation metrics.  }
		\label{table:ct_rgb}
	\end{table}
	
	\paragraph{Cross-Scale Pyramid Alignment}
	
	Different from existing works that apply temporal alignment on a specific scale or multiple scales individually, we propose a cross-scale pyramid alignment module (CSPA) that enables a more accurate alignment. As shown in Table~\ref{table:ablation}, the model with the proposed CSPA leads to a 0.66dB improvement on PSNR compared with the baseline. Furthermore, we also give some visual results for qualitative evaluation in Fig.~\ref{fig:abcspa}. The CSPA benefits our model in restoring the structure of the car and patterns on the clothes more clearly. 
	
	\paragraph{Temporal Consistency}
	\begin{figure}[t]
		\centering
		\includegraphics[width=1.0\columnwidth]{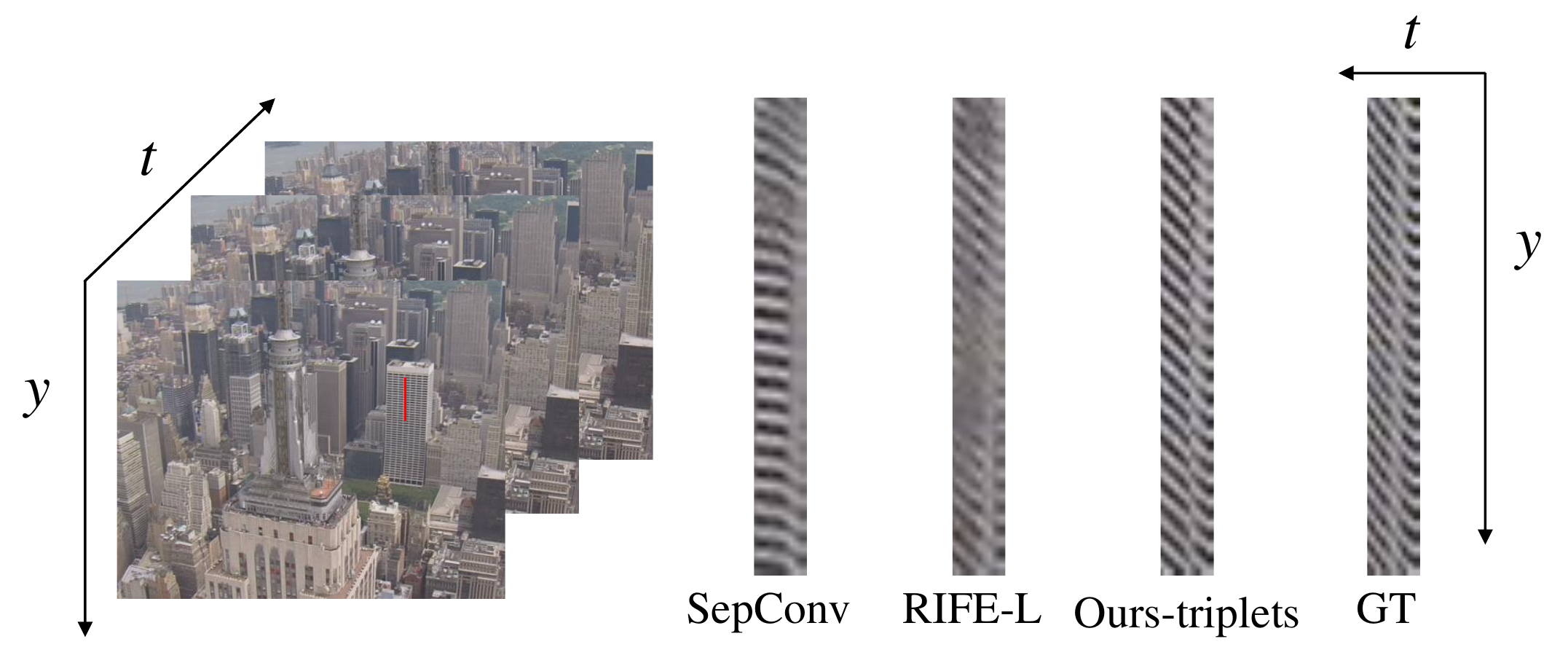} 
		
		\caption{Temporal consistency analysis of three single-frame VFI approaches. The SepConv~\cite{niklaus2017videosep} and RIFE-L~\cite{huang2020rife} cannot generate continuous signals, while our algorithm shows a smoother transition. The sample comes from Vid4~\cite{liu2013bayesian}.} 
		\label{fig:tsmooth}
	\end{figure} %
	Apart from the quantitative evaluation of PSNR and SSIM, temporal consistency~\cite{lai2018learning,zhang2020there} is also an important measure within the realm of video frame interpolation. We compare our method with two representative methods including SepConv~\cite{niklaus2017videosep} and RIFE-L~\cite{huang2020rife} in Fig.~\ref{fig:tsmooth}. It is observed that SepConv and RIFE-L generate blurry and inconsistent patterns along the time axis, while our method successfully restores the correct and consistent patterns compared with the ground truth.

	\vspace{-0.15in}
	\subsection{Extension}
	
	\paragraph{Video Frame Extrapolation}
	%
%
	\begin{table}[t]	
		\small
		\setlength{\tabcolsep}{3pt}
		
		\begin{center}
			\begin{tabular}{cccccccccc}
				\hline
				\multirow{2}{*}{Methods} & \# Parameters & \multicolumn{2}{c}{Vimeo-Triplets-Test}    &\multicolumn{2}{c}{Middlebury} \\ \cmidrule(r){3-4} \cmidrule(r){5-6}
				&  (Million)      & PSNR     &SSIM               & PSNR     &SSIM      		\\ \hline
		    
				SepConv~\cite{niklaus2017videosep}       &21.7               &30.42    &0.9170           & 32.21   &0.9546     		\\ 
				
				FLAVR~\cite{NEURIPS2019_d045c59a}       &{42.1}                 &31.14    &0.9268              & 32.90    &0.9619                    		  \\ \hline
				Ours-extra.                                   &22.4                 &{\bf 32.05}     &{\bf0.9395}              &{\bf34.66}   &{\bf 0.9765}            \\ \hline
			
			\end{tabular}
		\end{center}
		\caption{Quantitative comparison of SepConv, FLAVR, and our method for video frame extrapolation. The models are trained on the Vimeo-Triplets-Train dataset and evaluated on the Vimeo-Triplets-Test set and Middlebury benchmarks. The best PSNR(dB)/SSIM results are highlighted in {\bf bold}.}
		\label{fig:extrapolation}
		\vspace{-8pt}
	\end{table}
	\begin{figure*}[t]
		\centering
		\includegraphics[width=1.0\linewidth]{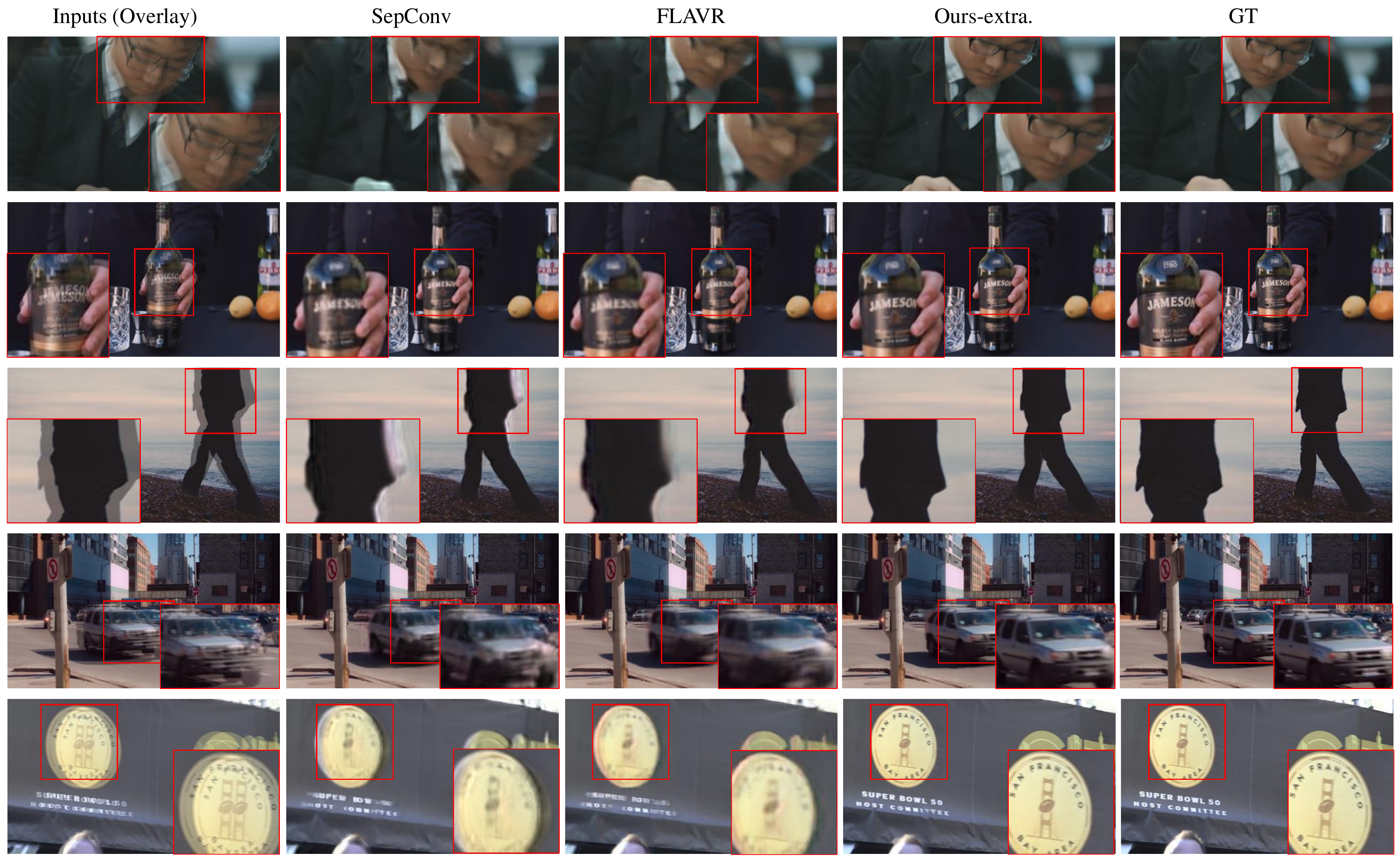} 
		
		\caption{Visual comparison of our method and two recent methods for video frame extrapolation.} 
		\vspace{-0.15in}
		\label{fig:ext_sota}
	\end{figure*} %

	We make an exploration to extend our method to the video frame extrapolation task. Unlike video frame interpolation, extrapolation aims to synthesize the future frames based on the observed historical frames. All the flow-based methods are heavily dependent on the pre-defined displacements, making them unsuitable for extrapolation. In this case, we compare our method with SepConv~\cite{niklaus2017videosep} and FLAVR~\cite{kalluri2020flavr} as they do not require optical-flow information. SepConv adopts a U-Net to regress a pair of separable 1D kernels to perform convolutional operations on the two input frames. FLAVR proposes an efficient 3D convolutional neural network for reconstruction. For a fair comparison, we retrain the three models from scratch under the same experimental setting on Vimeo-Triplets. More specifically, we need to predict a future frame $I_3$ from historical $\{ I_1,I_2\}$. We use fewer residual blocks~(18 residual blocks) in the reconstruction module, which makes our model has comparable parameters~(22.4M) with the SepConv~(21.7M) and much fewer parameters than FLAVR~(42.1M).
	
	As shown in Table~\ref{fig:extrapolation}, compared with SepConv and FLAVR, our model boosts PSNR by 1.63dB and 0.91dB, respectively. In addition, our model is nearly {\bf 2} times smaller than the FLAVR. As depicted in Fig.~\ref{fig:ext_sota}, our method produces sharper edges and fewer artifacts. Especially in the last image, our model can produce recognizable characters. In a nutshell, both the quantitative and qualitative results demonstrate that our model is capable of generating high-quality extrapolated frames.

	\paragraph{Synthetic Frames for Video Super-Resolution}
	
\begin{figure}[ht]
		\centering
		\includegraphics[width=1.0\columnwidth]{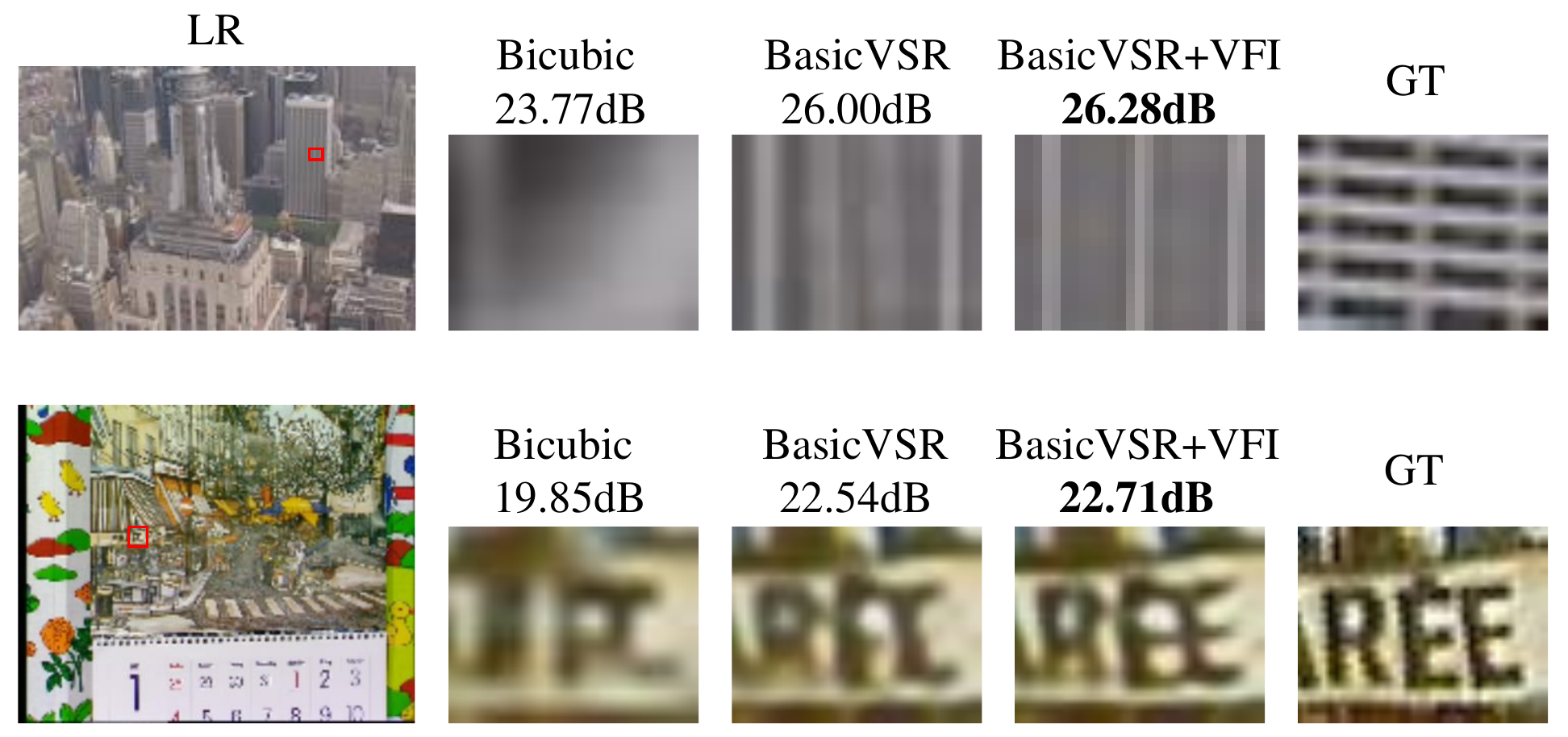} 
		
		\caption{Visual results of BasicVSR and ``BasicVSR + VFI" for the video super-resolution. } 
		\label{fig:vsrviz}
		
	\end{figure}

	We illustrate that the interpolated frames by our model also benefit video super-resolution methods. A well-trained recurrent framework BasicVSR~\cite{chan2021basicvsr} is adopted due to its strong performance as well as the flexibility to allow an arbitrary number of input frames. In the basic setting, we take three consecutive low-resolution frames as input:
	\begin{equation}
		\tilde{I}_0^{HR} = \mathop{f_{vsr}}(I_{-1}^{LR},I_{0}^{LR},I_{+1}^{LR}),
	\end{equation}
	Then, we utilize our model to generate two intermediate low-resolution frames $I_{-0.5}^{LR}$ and $I_{+0.5}^{LR}$ from $\{I_{-1}^{LR},I_{0}^{LR}\}$ and $\{I_{0}^{LR},I_{+1}^{LR}\}$, respectively. With the same BasicVSR model, the high-resolution image is reconstructed as
	\begin{equation}
		\tilde{I}_0^{HR} = \mathop{f_{vsr}}(I_{-1}^{LR},I_{-0.5}^{LR},I_{0}^{LR},I_{+0.5}^{LR},I_{+1}^{LR}).
	\end{equation}
	The quantitative comparison is shown in Table~\ref{table:vsrvis}. It suggests that the two additional frames generated by our model can boost the performance of BasicVSR by 0.1dB and 0.012 on PSNR and SSIM. Besides, Fig.~\ref{fig:vsrviz} shows that the results with more input frames~(denoted as ``BasicVSR + VFI") have more plausible structures. The experiment clarifies that the high-quality interpolated frames obtained from our VFI model are beneficial for the VSR method.
	
	\begin{table}[t]
		
		\centering
		\small
		\begin{tabular}{c|c|c } 
			\hline
			Method                        &PSNR~(dB)        &SSIM         \\ \hline
			
			BasicVSR                & 26.65  &0.7545            \\ 
			BasicVSR+VFI             & 26.75 ({\color{red} $\uparrow$ 0.1}) &0.7620 ({\color{red} $\uparrow$0.075})          \\  \hline
		\end{tabular}
		\caption{The quantitative analysis of our VFI method for the video super-resolution task.  }
		\vspace{-0.15in}
		\label{table:vsrvis}
	\end{table}

	\subsection{Natural Phenomena}
	\label{sec:np}

\begin{figure}[ht]
		\centering
		\includegraphics[width=1.0\columnwidth]{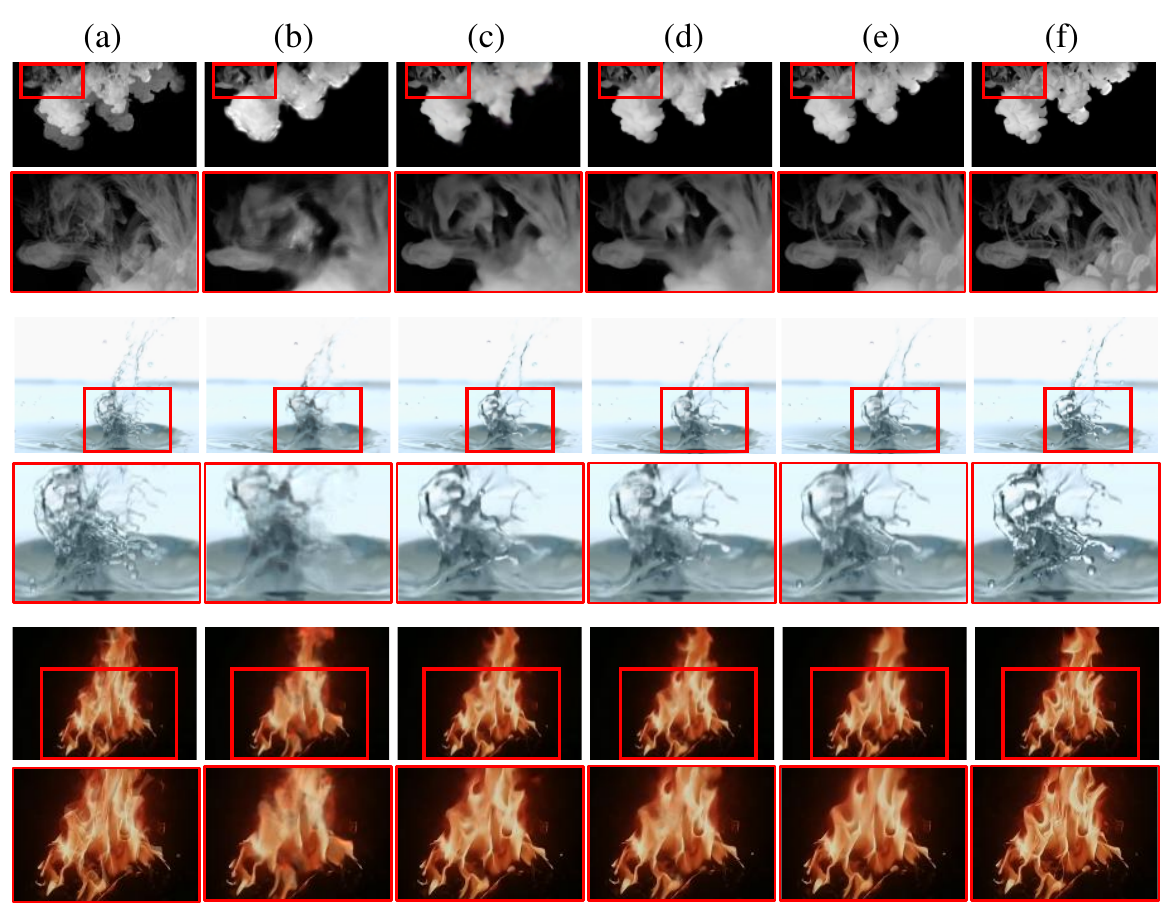} 
		
		\caption{Visual results for natural phenomena. (a) refers to the overlayed input frames. (b-f) represent the results of SepConv~\cite{niklaus2017videosep}, FeFlow~\cite{gui2020featureflow}, RIFE-L~\cite{huang2020rife},  ours-triplets and groundtruth.} 
		\label{fig:np}
		
	\end{figure}

	In this paper, following the most widely used setting in~\cite{xue2019video,niklaus2017videosep,niklaus2020softmax,niklaus2020softmax,park2020bmbc,cheng2020video,bao2019depth,choi2020channel,niklaus2021revisiting,cheng2021multiple,gui2020featureflow,chen2021pdwn,bao2019memc,huang2020rife}, we also focus on studying the problem of frame interpolation in general scenes. Taking a further step, we evaluate the generalization ability of our model (trained on the Vimeo-Triplets) by transferring it to out-of-domain natural scenarios. To this end, we first collect over 100 video clips from the Internet~(we will make this newly collected dataset publicly available for research uses later), containing dynamic textures of natural phenomena. As shown in the Table~\ref{table:np}, our method outperforms others by a large margin, yielding a 1.37dB gain. Besides, from the visual examples in Fig.~\ref{fig:np}, we observe that our method recovers a better and more natural texture. It is noteworthy that, without any physical priors, e.g., the diffusion of smoke, nor fine-tuning on a specific texture class~\cite{danier2021texture}, our approach still produces relatively plausible contents in these examples. All these observations demonstrate the generalization ability of our method.

	\begin{table}[t]
		
		\centering
		\small
		\setlength{\tabcolsep}{3pt}
		\begin{tabular}{c|cccc } 
			\hline
			Method                        &SepConv~\cite{niklaus2017videosep}        &FeFlow~\cite{gui2020featureflow}  &RIFE-L~\cite{huang2020rife} &Ours-triplets        \\ \hline
			
			PSNR~(dB)               & 22.57  &32.33    &32.54 & {\bf 33.91}         \\ 
			SSIM             & 0.8216 & 0.9487 &0.9512 & {\bf0.9611}        \\  \hline
		\end{tabular}
		 \caption{The quantitative comparison on the collected cases of natural phenomena. }
		\vspace{-0.15in}
		\label{table:np}
	\end{table}

	\subsection{Limitations and Disscussions}
	\label{fc}
	
\begin{figure}[ht]
		\centering
		\includegraphics[width=1.0\columnwidth]{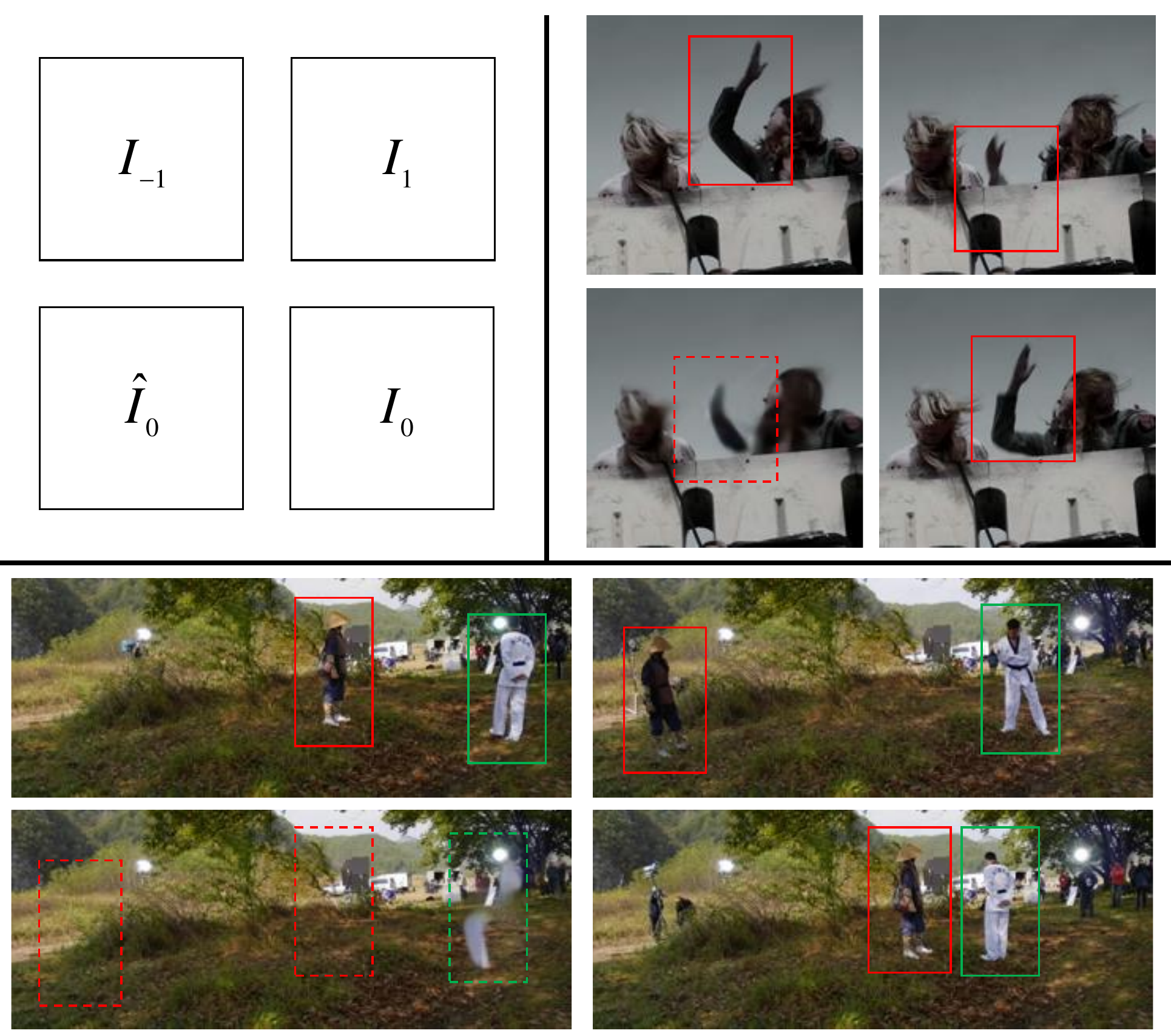} 
		
		\caption{Two failure cases of our method. } 
		\label{fig:fc}
	\end{figure}

	Though our model can generate high-quality results for most cases, it still faces a challenge for some complicated samples. We show two failure cases of our methods in Fig.~\ref{fig:fc}. For the first example, the woman's right arm is occluded by the wall in frame $I_{1}$, making it difficult to estimate accurate correspondence. As a result, our method tends to generate overly smooth content. In terms of the second sample, the man in the red box has an extremely large motion and the man in the green box turns around. Our method fails to hallucinate the corresponding objects in the middle frame.
	\zk{These challenging scenarios often require specific knowledge, e.g., physics-based priors, to enhance the model capability. For example, skeletal constraints (if exploited well) could be strong clues for interpolating human bodies. This can be a potential direction for future study. }
	\vspace{-0.15in}
	\section{Conclusion}
	\label{sec:conclusion}
	We present a novel and effective video frame interpolation approach. The proposed texture consistency loss relaxes the strict contraint of the pre-defined ground truth and the cross-scale pyramid alignment is able to make better use of multi-scale information, making it possible to generate much clearer details. Comprehensive experiments have demonstrated the effectiveness of our method to interpolate high-quality frames. Also, we show that our model is easily tailored to the video extrapolation task. And our interpolated frames are proven to be useful in improving the performance of existing video super-resolution methods. In the future, we plan to study the potential of our method on other video restoration problems, such as video deblurring and video denoising.
	
	
	
	%
	


	%
	
	\ifCLASSOPTIONcaptionsoff
	\newpage
	\fi

	
	
	%
	
	
	\bibliographystyle{IEEEtran}
	\bibliography{VFI_ARXIV}
\begin{appendices}

\section{More Visual Comparison of TCL}
\begin{figure*}[h]
		\centering
		\begin{subfigure}{\linewidth}
			\includegraphics[width=1.0\linewidth]{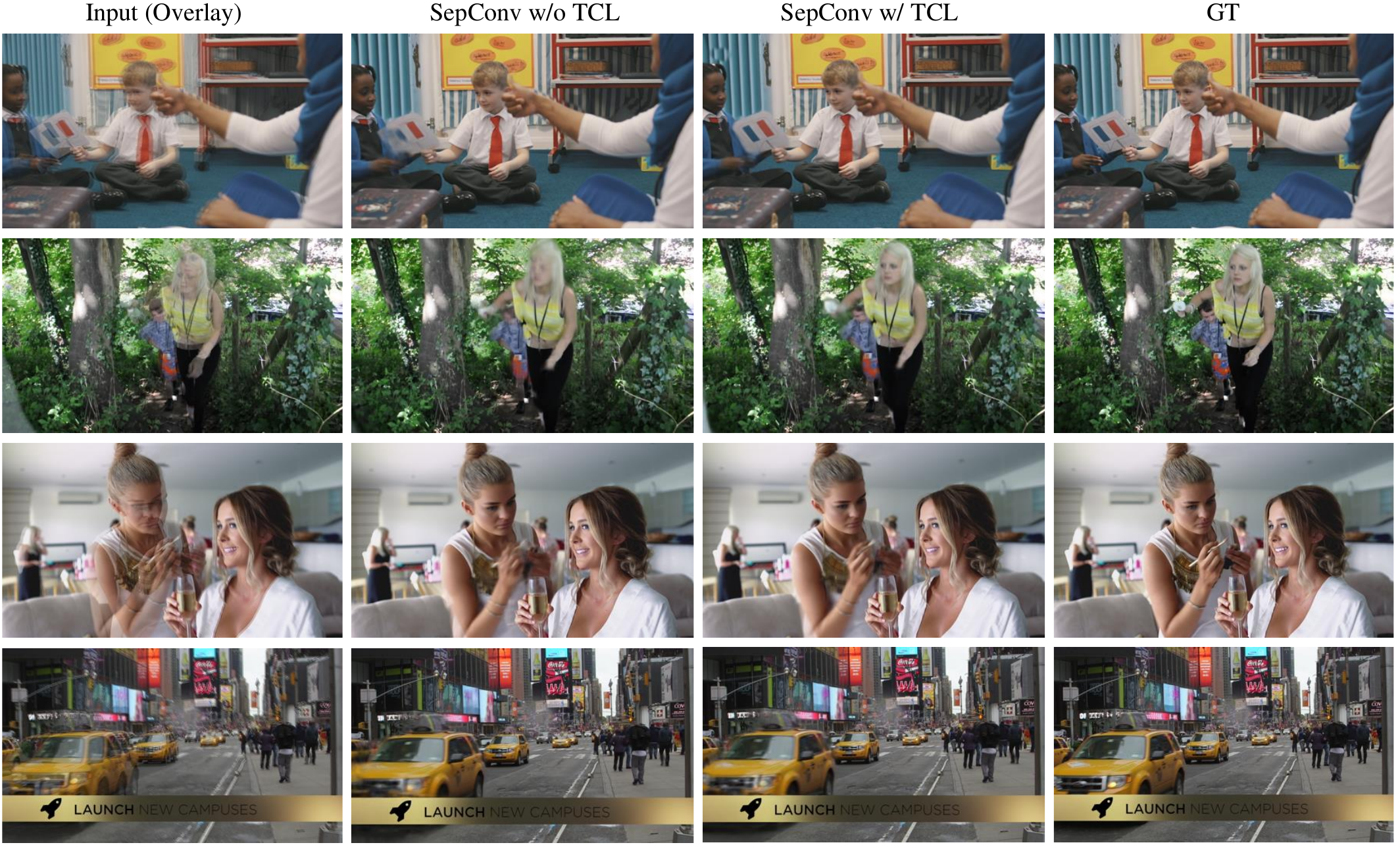} 
			
			\caption{Visualized results of the SepConv~\cite{niklaus2017videosep} without/with our TCL. } 
			\label{fig:append1}
			
		\end{subfigure}
		\hfill
		\begin{subfigure}{\linewidth}
			\includegraphics[width=1.0\linewidth]{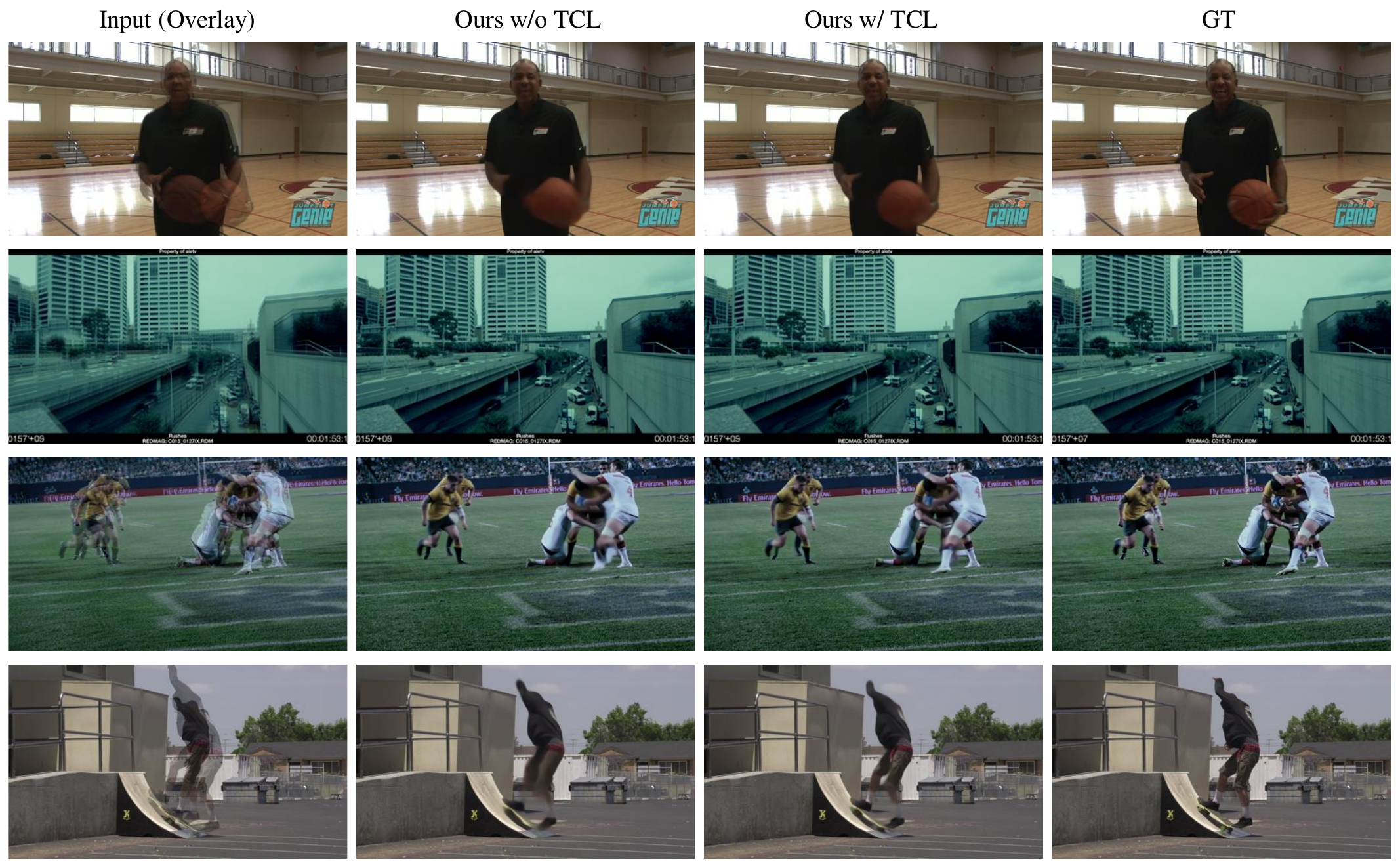} 
			
			\caption{Visualized results of our models trained without/with our TCL.}
			\label{fig:append2}
		\end{subfigure}
		
		\label{fig:pmimpacts}
	\end{figure*}

	As shown in Fig.~\ref{fig:append1} and Fig.~\ref{fig:append2}, we give more visual examples of SepConv~\cite{niklaus2017videosep} and our method with/without TCL. It is clear that the proposed TCL is benefical in hallucinating more plausible structures.
	\label{sec:AppenA}

	\section{Discussion of Model-{1,2,3} in Section~\ref{align}}
	\label{sec:AppenB}
	Except for the alignment module, the Model-{1,2,3} in Sec.~\ref{align} share the same framework architecture as shown in Fig.~\ref{fig:framework}. The Model-1 adopts a single-scale alignment of which computational complexity is
	\begin{equation}
		C_{Model-1} = k_1*N,
	\end{equation}
	where $N$ is the number of pixels on the highest resolution and $k_1$ denotes the number of operations applied on a single pixel. As for Model-2, the cost volume is obtained by calculating pixel-to-pixel correlations. Taking the alignment of highest-resolution features for example, apart from the current scale, it also requires to compute the cross-scale correlations with respect to the $1/2$ and $1/4$ resolution features, resulting in
	\begin{equation}
		C_{Model-2} = (1+1/4+1/16)*k_2*N^2,
	\end{equation}
	where $k_2$ represents the pixel-wise operations of Model-2. In contrast, the Model-3 (our CSPA) performs the cross-scale alignment with computational complexity as
	\begin{equation}
		C_{Model-3} = (1+1/4+1/16)*k_3*N,
	\end{equation}
	where $k_3$ counts the number of pixel-wise operations of Model-3. Considering $k_1$, $k_2$ and $k_3$ are much smaller than the pixel number $N$, our CSPA shares a comparable computational complexity of $O(N)$ with the Model-1, much smaller than Model-2 of $O(N^2)$.
	
	\end{appendices}
	\end{document}